\documentclass[pdflatex,sn-apa]{sn-jnl}

\usepackage{graphicx}
\usepackage{multirow}
\usepackage{amsmath,amssymb,amsfonts}
\usepackage{amsthm}
\usepackage[mathscr]{euscript}
\usepackage[title]{appendix}
\usepackage{xcolor}
\usepackage{textcomp}
\usepackage{manyfoot}
\usepackage{booktabs}
\usepackage{algorithm}
\usepackage{algorithmicx}
\usepackage{algpseudocode}
\usepackage{listings}

\usepackage{silence}
\theoremstyle{thmstyleone}

\theoremstyle{thmstyletwo}

\theoremstyle{thmstylethree}

\begin{document}

\title[Article Title]{From Prototypes to Sparse ECG Explanations: SHAP-Driven Counterfactuals for Multivariate Time-Series Multi-class Classification}

    \author*[1,3]{\fnm{Maciej} \sur{Mozolewski}}\email{m.mozolewski@uj.edu.pl}
    \equalcont{These authors contributed equally to this work.}

    \author[2]{\fnm{Betül} \sur{Bayrak}}\email{betul.bayrak@ntnu.no}
    \equalcont{These authors contributed equally to this work.}

    \author[2]{\fnm{Kerstin} \sur{Bach}}\email{kerstin.bach@ntnu.no}
    \author[3,4]{\fnm{Grzegorz~J.} \sur{Nalepa}}\email{grzegorz.j.nalepa@uj.edu.pl}

    \affil[1]{
      \orgdiv{Jagiellonian Human-Centered AI Lab, Mark Kac Center for Complex Systems Research},
      \orgname{Jagiellonian University},
      \orgaddress{\street{\L{}ojasiewicza 11}, \city{Krak\'ow}, \postcode{30-348}, \country{Poland}}
    }

    \affil[2]{\orgdiv{Department of Computer Science}, \orgname{Norwegian University of Science and Technology (NTNU)}, \orgaddress{\street{H\o gskoleringen 1}, \city{Trondheim}, \postcode{7034}, \country{Norway}}}

    \affil[3]{\orgdiv{Department of Human-Centered Artificial Intelligence}, \orgname{Institute of Applied      Computer Science, Jagiellonian University}, 
        \orgaddress{\street{\L{}ojasiewicza 11}, \city{Krak\'ow}, \postcode{30-348}, \country{Poland}}
    }

    \affil[4]{\orgdiv{School of Information Technology}, \orgname{Halmstad University}, \orgaddress{
    \city{Halmstad},
    \country{Sweden}}}

    \abstract{
        In eXplainable Artificial Intelligence (XAI), instance-based explanations for time series have gained increasing attention due to their potential for actionable and interpretable insights in domains such as healthcare. 
        Addressing the challenges of explainability of state-of-the-art models, we propose a prototype-driven framework for generating sparse counterfactual explanations tailored to 12-lead ECG classification models. 
        Our method employs SHAP-based thresholds to identify critical signal segments and convert them into interval rules, uses Dynamic Time Warping (DTW) and medoid clustering to extract representative prototypes, and aligns these prototypes to query R-peaks for coherence with the sample being explained. 
        The framework generates counterfactuals that modify only 78\% of the original signal while maintaining 81.3\% validity across all classes and achieving 43\% improvement in temporal stability. We evaluate three variants of our approach, Original, Sparse, and Aligned Sparse, with class-specific performance ranging from 98.9\% validity for myocardial infarction (MI) to challenges with hypertrophy (HYP) detection (13.2\%). This approach supports near real-time generation ($<$ 1 second) of clinically valid counterfactuals and provides a foundation for interactive explanation platforms. Our findings establish design principles for physiologically-aware counterfactual explanations in AI-based diagnosis systems and outline pathways toward user-controlled explanation interfaces for clinical deployment.

    }

    \keywords{Explainable AI, ECG explanations, Sparse time-series counterfactuals, Rule extraction}

    \maketitle

    \section{Introduction}\label{sec:introduction}
    \label{sec:intro}

One of the key challenges in healthcare engineering is the automatic analysis of ECG signals for early detection of cardiac abnormalities and timely clinical intervention. 
Despite digital recording, interpretation of 12‐lead ECGs still depends on expert visual assessment of waveforms and remains challenging due to inter‐individual variability in gender, electrode placement, and electrophysiological properties~\citep{VANDAM202167}. 
Conversely, recent innovations in healthcare have made the ECG increasingly a commodity, while training in ECG interpretation remains limited, placing a greater workload on experts and potentially hampering the early detection of cardiac patients~\citep{PRONIEWSKA202345}. 

On the other hand, the availability of large‐scale, clinically accurate 12‐lead ECG datasets (e.g., PTB‐XL~\citep{wagner_2020_ptbxl,wagner_2022_ptbxl} and PTB‐XL+~\citep{Strodthoff2023} has enabled the development of high‐performance deep learning classifiers capable of detecting multiple cardiac conditions within a single recording. 
However, the inherent complexity of these models renders their decision‐making processes opaque to end users~\citep{Holzinger2019}, underscoring the need for tailored explainable AI techniques in ECG interpretation.
Crucially, the challenge extends beyond the algorithm itself. 
As~\cite{SCHMITT2023100520} argues about safety-critical digital ecosystems, effective deployment requires seamless integration, real-time responsiveness, and strict governance. 
Our framework aligns with these principles by providing efficient and verifiable explanations.
Those requirements are equally vital for clinical decision-support systems.
Consequently, there is a pressing need for tailored explainable AI techniques that not only clarify model outputs but also fit within rigorous healthcare workflows. 

Despite progress in SHAP-based local explanations~\citep{Lundberg2017} such as WindowSHAP~\citep{WindowSHAP2023}, current methods might be too slow for real-time deployment~\citep{vanlooveren2020interpretablecounterfactualexplanationsguided}. 
Counterfactual generation techniques frequently violate ECG morphology and temporal coherence, yielding physiologically implausible waveforms. 
~\cite{baig2025arrhythmiavisionresourceconsciousdeeplearning} describe a lightweight ECG analysis system tailored for wearable and embedded platforms. 
They highlight the demand for efficient models capable of real-time analysis in resource-limited settings, yet many XAI techniques still introduce substantial computational overhead. 
This persistent shortfall in explainability performance motivates our work: our framework delivers potentially low-latency inference, actionable counterfactual explanations, which are physiologically faithful as they are grounded in real patient prototypes, and the potential for point-of-care deployment on standard clinical workstations.

In recent works, explainable AI methods have been applied to ECG interpretation with promising results. 
Image-based techniques, such as those of~\cite{image_based_ecg_2023}, overlay heatmaps on 12-lead ECG images to highlight diagnostically relevant waveform features and improve model interpretability. 
They further advocate for advanced XAI strategies: counterfactual explanations and rule-based methods, to boost both transparency and robustness.  
Similarly,~\cite{TANYEL2025107227} employed DiCE-based counterfactuals~\cite{mothilal2020dice} on the PTB-XL dataset, validating generated examples with two cardiologists and demonstrating their clinical relevance.  
Building on these complementary approaches, our framework combines rule-based explanations, heatmaps and efficient counterfactual generation to deliver real-time, actionable insights.

In the context of ECG counterfactual explanations, \emph{sparsity} refers to the proportion of the signal that requires modification to achieve a class transition. A sparse counterfactual modifies only a minimal subset of time points across the 12 leads, rather than perturbing the entire recording. This sparsity is crucial for clinical interpretability since sparse explanations highlight specific waveform segments that drive diagnostic changes instead of presenting wholesale signal replacements. By focusing clinical attention on these critical regions, which states a challenge in the literature, sparse counterfactuals align with how cardiologists naturally interpret ECGs, examining specific morphological features rather than global waveform differences.

   \begin{figure}[h!]
        \centering
        \includegraphics[width=\textwidth]{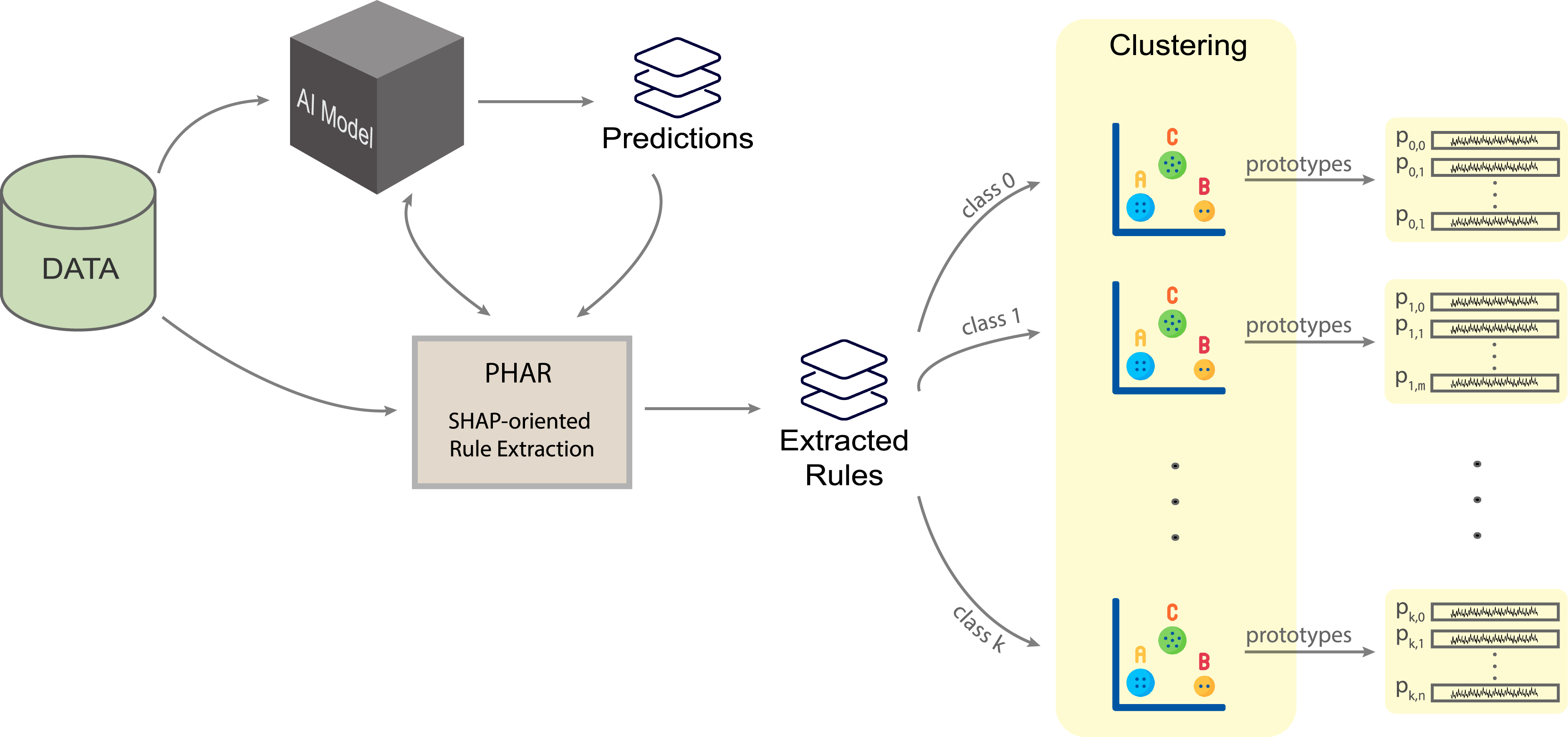}
        \caption{Overview of the proposed framework. ECG data is classified by an AI model, and SHAP-based rule extraction identifies important features. Rules are clustered per class to select prototype ECGs. For a query, the system retrieves prototypes, applies R-peak alignment and sparsity optimization, and generates counterfactual explanations.}
        \label{fig:proto}  
    \end{figure}
Our framework addresses these challenges by combining SHAP-derived rule extraction and prototype-based sparse counterfactual generation (see Figure~\ref{fig:proto} for an overview). 
First, employing the Post-hoc Attribution Rule extraction (PHAR) framework~\citep{mozolewski2025phar}, we compute SHAP values and select features above a global threshold, forming concise interval rules. 
Next, we measure distances between rule-defined series using Dynamic Time Warping (DTW)~\citep{DWT1163055} and employ clustering followed by medoid selection to extract representative prototypes for each class.
Finally, we apply our optimisation step, which consists of aligning and scaling these prototypes to the query ECG by matching characteristic R-peaks. 
To enrich comprehensibility, we interviewed clinicians, including cardiology and electrophysiology experts, using semi-structured, one-on-one sessions. 
Participants reviewed lead-wise importance charts, signal overlays, and both single- and multi-overlay counterfactual examples annotated with interval rules. 
They rated readability, design clarity, and information density, and offered improvement suggestions. 

In summary, our method delivers fast, physiologically valid counterfactual ECG explanations. 
It combines SHAP-derived interval rules with precomputed prototypes to preserve both waveform morphology and global rhythm. 
By precomputing prototypes, counterfactuals are generated in near real time with the cost of querying and optimization. 
Preliminary clinician interviews provided initial positive feedback on the readability, clarity, and information density of our visualizations.
This approach strengthens transparency and trust in AI-driven ECG analysis.

The main contributions of this work are as follows:
\begin{itemize}
  \item A novel counterfactual generation approach that proposes sparse counterfactuals from prototypes derived from existing data samples, with optimization applied. 
  \item A prototype-based strategy enabling real-time application by using pre-computed instances combined with an optimization process, resulting in tailored explanations with reduced computational cost.
  \item A quantitative evaluation framework specialized on ECG counterfactual explanations that covers validity, sparsity, stability, and margin metrics.
  \item A visualization framework integrating SHAP explanations with single- and multiple-counterfactual overlays, qualitatively evaluated with three cardiology experts.
\end{itemize}

The remainder of this article is structured as follows.  
Section~\ref{sec:background} reviews the background and related work.  
Section~\ref{sec:method} describes our proposed method for generating counterfactual explanations of ECG signals, including adaptation of the \emph{Post-hoc Attribution Rule extraction (PHAR)} framework (Section~\ref{sec:rule-xai}), prototype selection (Section~\ref{sec:prototypes}), and the explanation generation process (Section~\ref{sec:explanation}).  
Section~\ref{sec:experiments} presents the experimental evaluation, including quantitative results and qualitative interviews with ECG analysts.  
Section~\ref{sec:discussion} discusses the advantages and limitations of our approach and outlines directions for future work.  
Finally, Section~\ref{sec:conclusion} concludes the article by summarizing the main contributions. 
Additional details are provided in the appendices: 
Appendix~\ref{secA3} describes the details of ECG model training and calibration,
Appendix~\ref{secA1} reports the metrics for ECG counterfactual evaluation,  
Appendix~\ref{secA2} presents example ECG counterfactual visualizations from the proposed method, and 
Appendix~\ref{secA4} outlines the expert evaluation protocol.

    \section{Background}\label{sec:background}
    \label{sec:bg}
\subsection{Explainable AI in Healthcare}
Recent work by~\cite{cliniciansvoice_2025} emphasizes the growing adoption of AI and XAI systems in healthcare. 
In the study's semi-structured interviews, clinicians prioritized feature-importance scores due to their immediate visual interpretability, which allows practitioners to identify key predictive factors without requiring deep knowledge of underlying algorithms like SHAP~\citep{Lundberg2017} or LIME~\citep{LIME}.
However, respondents cautioned that this abstract simplicity can be deceptive, provoking skepticism regarding the validity of reducing complex clinical scenarios to basic additive factors. 

In the specialized domain of clinical decision support, SHAP and LIME have emerged as the most prominent paradigms for interpreting complex medical predictions~\citep{Sannidhi2022AutonomousDiseasePredictions, Ahmed2025DiabetesXAI}. 
A comprehensive literature review conducted by~\cite{Manimaran2025ECGLiteratureReview} reveals that SHAP is the most frequent choice for electrocardiogram (ECG) analytics, appearing in 21.9\% of studies while LIME follows at 9.4\%~\citep{Manimaran2025ECGLiteratureReview}. 
While both frameworks facilitate transparency, they utilize fundamentally different mechanisms since SHAP is rooted in additive feature attribution from game theory and LIME focuses on local feature perturbation~\citep{Salih2025ShapLime}. 
However, as highlighted by Salih et al., post-hoc attribution methods like SHAP and LIME often struggle with highly correlated biomedical signals, which may obscure true physiological relationships. 
This inherent instability, as noted by~\cite{Singh2025Beyond}, poses a significant risk for misleading clinical interpretations if raw importance scores are used in isolation. 

As demonstrated in the clinical interviews conducted by~\cite{cliniciansvoice_2025}, alternative paradigms such as counterfactual explanations and rule-based models were acknowledged for offering deeper insights or mirroring human reasoning, but faced practical criticism. 
Specifically, counterfactuals were viewed as potentially causally inconsistent or irrelevant due to their focus on hypothetical patients, whereas rule-based systems were deemed too cumbersome for rapid decision-making. 

Holzinger et al.~\citep{Holzinger2019Causability} argue that effective XAI in medicine requires intuitive explanation interfaces and tight human‐in‐the‐loop integration to ensure clinical usability. 
They emphasize the trade‐off between ante‐hoc models, which offer built‐in transparency but may sacrifice accuracy, and post‐hoc methods that provide flexible explanations for black‐box predictors. 
Moreover, they advocate for causal frameworks capable of generating counterfactual insights, thereby aligning technical explainability with clinicians’ reasoning. 

The relevance of XAI in clinical decision-making is further underscored by a systematic review by~\cite{Antoniadi2021Systematic}, which identifies that the core value of interpretability in Clinical Decision Support Systems (CDSS) is to provide local explanations that support the decision-making process of medical practitioners' in specific cases. 
They observe that XAI research is mainly focused on tabular datasets, while other modalities are less frequent. 
Crucially, the authors highlight that ensuring adoption, as opposed to simply technical acceptance, requires bridging the lack of user-centric validation through expert-engaged studies. 

This ongoing challenge of validation is echoed by~\cite{Salih2024EvalXAI}, who review 213 studies on XAI in cardiology and report that 43\% of works do not validate explanations, 37\% rely solely on literature comparison, 11\% involve clinical experts, and 11\% apply statistical or proxy-grounded evaluation. 
Commonly used techniques for ECG interpretation include SHAP (global and local importance scores) and Grad-CAM (activation maps), sometimes augmented with LIME. 
In the minority of studies engaging cardiologists, evaluation focuses on the alignment between model-highlighted critical points and expert interpretation. 
The authors emphasize the necessity of application-grounded validation with cardiologists, the adoption of proxy-grounded methods (which assess explanation fidelity by removing attributed features and retraining the model), permutation tests (which quantify the effect of feature shuffling on predictions) and a hybrid evaluation strategy combining human-, proxy-, and literature-grounded approaches to ensure clinical relevance and technical reliability. 

A similar comprehensive evaluation approach is adopted by~\cite{darias2024empirical} and~\cite{darias2025evaluating}, who employ addition and deletion metrics that systematically assess explanation fidelity by removing or adding attributed features and measuring the resulting impact on model predictions. 
These proxy-grounded methods provide quantitative measures of explanation quality by evaluating how feature modifications affect classification performance, complementing human-centered evaluation approaches. 
Such multi-faceted evaluation frameworks are essential for establishing the reliability and clinical utility of XAI systems in healthcare applications.

The necessity of such multi-faceted frameworks is further underscored by~\citep{Amann2022ToExplain}, who argue that the value of explainability in CDSS is inherently context-dependent. 
They emphasize that XAI should not merely provide data, but facilitate ``meaning-making'' for the clinician by balancing technical feasibility with the user's cognitive constraints. 
This perspective provides a principled foundation for our methodology. 
Specifically, our sparsity optimization directly addresses the requirement to minimize information overload, ensuring that only the most critical diagnostic changes are presented. 
Furthermore, the use of R-peak alignment and regional highlighting ensures that the explanations align with clinical intuition and facilitate rapid interpretation. 
By adhering to these design principles, our system moves beyond raw feature attribution toward a task-oriented CDSS that supports actionable decision-making.

\subsection{Interactive Visualization for Expert Feedback}
Interactive machine learning (iML) highlights the critical role of end‐user engagement through iterative model refinement. 
Wondimu et al.~\citep{Wondimu2022InteractiveML} demonstrate that interactive visualizations, such as multi‐view decision path diagrams and activation heatmaps, empower users to inspect model behavior, adjust parameters in real time, and directly observe the impact of their inputs. 
The authors emphasize that transparency, user control, and ongoing human–machine collaboration are essential foundations for trustworthy AI in high‐stakes domains.

Theissler et al.~\citep{xai_ts_class_2022} propose a four‐fold taxonomy of XAI for time‐series: time‐point attribution, subsequence methods, instance‐based strategies (including counterfactuals), and rule/logic frameworks. 
They observe that ECG applications predominantly rely on attribution heatmaps and subsequent highlighting, yet often lack interactive evaluation and standardized benchmarks. 
The authors emphasize visualization (saliency maps, activation heatmaps, annotated overlays) as essential for translating model decisions into experts' insights. 
Key research gaps include the development of higher‐order and truly model‐agnostic explanations, domain‐specific interpretability techniques, user‐friendly interfaces, and advanced evaluation protocols. 

\subsection{Deep Learning for ECG Classification}\label{sec:DL_for_ECG}
Deep neural networks achieve high accuracy on multivariate time series tasks~\citep{Wang2024}.
Classification models have shown state-of-the-art performance.
An exemplary deep learning model for ECG analysis is described by~\cite{Hannun2019}, which achieves cardiologist‐level arrhythmia detection but does not incorporate any explainability (XAI) techniques.
~\cite{baig2025arrhythmiavisionresourceconsciousdeeplearning} develop lightweight 1D CNN models
for on‐device arrhythmia classification and integrate popular XAI methods: Grad‐CAM to produce local activation maps that highlight key signal components (e.g., QRS complexes, T waves) and SHAP to assign per‐timepoint attribution scores across different diagnoses.

Furthermore, beyond standard CNN architectures, deep generative models have become integral to physiological signal analysis.
A recent systematic review by~\cite{NEIFAR2025103127} confirms their extensive application in ECG processing, identifying Generative Adversarial Networks (GANs) and Variational Autoencoders (VAEs) as the predominant architectures, primarily used for data augmentation and signal denoising.
However, in the context of resource-constrained environments, flow-based approaches offer distinct advantages.
\cite{Ibrahim2025EnhancingEC} conducted a comprehensive comparison of unsupervised anomaly detection filters, including Deep SVDD, Autoencoders, VAEs, and diffusion models.
Their study demonstrated that Normalizing Flows (NF) achieve a superior trade-off between computational efficiency and detection performance, often outperforming reconstruction-based approaches.
This capacity to accurately model complex probability densities makes NF a compelling choice for robust feature extraction in lightweight clinical systems.

\subsection{Counterfactual Explanations}
While standard CNN classifiers and attribution-based XAI techniques yield valuable insights, their raw visualizations can be cluttered and insufficiently streamlined for fast clinical interpretation.

To overcome the visual complexity of raw attribution maps, instance-based explanations provide intuitive alternatives by showing how an input would need to change to achieve a different outcome.
~\cite{Delaney2021InstanceBasedCounterfactual} introduces Native Guide, an instance‐based method for obtaining counterfactual explanations in time‐series classification. 
The approach retrieves the nearest opposite‐class exemplar from a reference database and adapts its key subsequences using classifier‐derived weight vectors, yielding counterfactuals that align closely with the original data distribution. 
The authors validate their method on multiple UCR benchmarks, including ECG200\footnote{\url{https://www.timeseriesclassification.com/description.php?Dataset=ECG200}}~\citep{Olszewski2001}, which contains pre‐segmented single‐beat electrocardiogram (ECG) signals. 
These simplified datasets are less complex than real‐world clinical recordings. 
They report superior performance over gradient‐based counterfactual techniques in terms of proximity, sparsity, plausibility, and diversity. 

Work by~\cite{Karlsson2020Tweaking} formalized the problem of ``time series tweaking'', distinguishing between local strategies (modifying specific shapelets) and global strategies (using nearest neighbors). 
In the local setting, they leverage Random Shapelet Forests (RSF) to detect discriminative subsequences, subsequently modifying the query signal to match the shapelet values required for a class change. 
Conversely, their global strategy utilizes a $k$-Nearest Neighbor ($k$-NN) approach to retrieve an existing instance from the target class, positing that historical examples offer inherently plausible explanations. 
While their global $k$-NN approach effectively identifies plausible counterfactuals in general time-series domains, its direct application to ECG analysis is limited by the lack of temporal alignment, as phase shifts (e.g., R-peak offsets) can distort the morphological comparison essential for cardiac diagnosis.

Positioning their work against these instance-based baselines, specifically comparing performance with~\cite{Delaney2021InstanceBasedCounterfactual} and the shapelet approaches of~\cite{Karlsson2020Tweaking},~\cite{Wang2024Glacier} proposed {Glacier}, a framework generating counterfactuals via gradient search within either the original feature space or an auto-encoder's latent representation. 
They report superior proximity and compactness compared to the aforementioned retrieval methods. 
A central contribution of their work is the integration of ``local constraints'' to guide the gradient descent, aiming to restrict modifications to specific time segments and suppress high-frequency noise. 
However, despite these regularizations, the method remains fundamentally reliant on numerical perturbation. 
Such mathematical deformation, while optimizing for decision boundaries, lacks explicit biological constraints, risking the introduction of signal artifacts that may not align with cardiac electrophysiology.

The potential of counterfactuals for clinical decision support has been demonstrated in the context of ECG analysis for datasets more closely related to clinical practice.
~ \cite{TANYEL2025107227} introduced the VCCE method for the detection of myocardial infarction (MI) using the PTB-XL dataset.
By utilizing the DiCE framework, they generate counterfactual clues through feature perturbations, providing clinicians with visual evidence of diagnostic transitions.
While their work highlights the high clinical alignment of optimization-based counterfactuals in binary classification, such approaches can be computationally intensive and may lack explicit rhythm consistency.
In contrast, our framework extends beyond binary MI detection to a multi-class setting and addresses the need for physiologically-consistent rhythm alignment and real-time inference.

While counterfactual and rule-based explanations offer deeper contextual understanding, they demand additional time and cognitive effort to assess. 
Successful deployment depends on so-called \textit{unremarkable AI} solutions that integrate seamlessly into existing workflows \citep{Yang2019Unremarkable} and on multidisciplinary collaboration between developers, researchers, and healthcare professionals.

\subsection{Prototype-based Explanations}
Prototypes, serving as class representatives to guide the search for counterfactual explanations, were discussed by \cite{vanlooveren2020interpretablecounterfactualexplanationsguided}. 
They introduced a model-agnostic method for generating counterfactual explanations in which the search is accelerated and interpretability is enhanced by guiding optimization with class prototypes.
In their approach, prototypes are derived either as latent-space centroids from an encoder or via class-specific k-d tree medoids, effectively replacing costly gradient-based searches. 
While their work focused on static benchmarks like MNIST or the Adult Census dataset, it established the foundation for using exemplars to improve both speed and interpretability.

Extending the concept of prototype guidance to the temporal domain,~\cite{Li2025Reliable} recently proposed a framework utilizing ShapeDBA.
This technique employs Dynamic Time Warping (DTW) on local structural features to compute a representative average (barycenter) that minimizes temporal distortion.
However, their validation relied on the TwoLeadECG dataset from the UCR Archive, which typically consists of short, pre-segmented heartbeats.
In contrast, clinical 12-lead recordings involve continuous signals with variable rhythms and multiple cardiac cycles.
While ShapeDBA mitigates alignment issues in short segments, our framework employs real patient medoids and explicit R-peak alignment to strictly preserve physiological fidelity across complex recordings, avoiding the potential smoothing artifacts of barycentric averaging.

Furthermore, effective prototype-based reasoning requires not just identifying a similar example, but explaining \textit{why} it is relevant.
~\cite{Karolczak2025} argue that users must understand which specific features make a prototype similar to the query instance.
They introduced the concept of ``alike parts'', which are segments identified by integrating feature importance scores from model-agnostic methods directly into the prototype selection process.
By demonstrating that fusing such importance scores, derived from various attribution techniques like SHAP or LIME, with example-based explanations significantly enhances user understanding and perceived transparency, their work provides a strong rationale for our approach.
This validates our strategy of utilizing importance-driven highlights to identify specific ECG regions, thereby streamlining the clinical reasoning process and ensuring that diagnostic waveform alterations are immediately apparent to the practitioner.

Synthesizing these perspectives, the literature establishes that seamless workflow integration and interactive visualization are essential for translating the high performance of deep learning models into clinical practice. 
However, a critical gap persists in generating explanations that are both physiologically plausible and rapidly interpretable. 
While conventional counterfactual methods often sacrifice biological realism to minimize the numerical distance to the query instance, and existing prototype-based approaches lack the precise temporal alignment required for complex ECG rhythms, our framework addresses these limitations.
By combining real-patient prototypes with explicit R-peak synchronization and sparsity optimization, we propose a solution that is physiologically consistent, visually intuitive, and designed to facilitate clinical integration and support diagnostic workflows. 

    \section{From SHAP-derived Rules to Sparse ECG Explanations}\label{sec:method}
        Our proposed method integrates rule-based feature selection with prototype-guided counterfactual generation to provide interpretable and clinically plausible explanations for ECG classification. 
    As illustrated in Figure~\ref{fig:proto}, the overall pipeline consists of three main components: 
    (i) extraction of concise interval rules from SHAP importance values with \emph{Post-hoc Attribution Rule extraction}~\citep{mozolewski2025phar}, 
    (ii) prototype selection via medoid clustering in DTW-based Multi-Dimensional Scaling (MDS) space, and 
    (iii) an explanation module that aligns selected prototypes with the input signal and generates sparse counterfactuals. 
    This modular structure enables fast and physiologically coherent explanations suitable for point-of-care use.

\subsection{PHAR: Rule Extraction from Feature Attributions}\label{sec:rule-xai}
    We use \emph{Post-hoc Attribution Rule extraction (PHAR)}, a model-agnostic framework developed by~\cite{mozolewski2025phar} for converting numeric feature attributions into concise, instance-specific rules.  
    \emph{PHAR} supports numeric attributions from methods like SHAP~\citep{Lundberg2017} or LIME~\citep{LIME}. 
    These attributions serve only as a preliminary filter to identify regions of interest for PHAR. 
    First, we select top features via a global threshold of the absolute attributions at a chosen percentile. 
    Then we apply controlled perturbations within each feature’s variance, identifying stable value intervals that preserve the original prediction. 
    While the framework is explainer-agnostic, we adopt SHAP as the importance source due to its solid game-theoretic foundations. 
    We prioritize this approach because it ensures that the initial attribution is theoretically distinct from the perturbation-based logic used in PHAR’s subsequent interval-identification step. 
    Any potential noise in the underlying explainer is mitigated by PHAR's internal consistency checks, which require prediction invariance within the identified intervals.
    Ultimately, the transition from abstract rule-based importance to explanations anchored in real-patient medoids (Section~\ref{sec:prototypes}) ensures that the final insights are not only stable but also physiologically plausible.

    The generated rules may facilitate transparent decision-making and support domain experts in validating model behavior, and can be visualized on time series as semi-factual scenarios~\citep{semi_ijcai2023p732}. 
    In this work, however, the rules from \emph{PHAR} serve to determine the most important features in ECG signals and are subsequently used for clustering. 
    In Section~\ref{sec:prototypes}, these rules are used to generate prototypes by clustering the rule sets for each class and selecting medoids, where the ECG instance corresponding to a medoid rule becomes the prototype and serves as a counterfactual. 
    To ensure this paper is self-contained, Section~\ref{sec:numeric_to_rules} provides a comprehensive overview of the core \emph{PHAR} mechanism as adapted for our ECG classification framework, while further details on rule fusion strategies can be found in~\cite{mozolewski2025phar}.  

    \subsubsection{Conversion of Numeric Importance Values to Rules} \label{sec:numeric_to_rules}
    This procedure, adapted from the \emph{PHAR} framework originally, concerns only the transformation of numeric feature attributions into instance-specific rules. 

    We have a set of \(N\) time series instances \(\{X_n\}\) with model output vectors \(\mathbf{y}_n\). We use each \(\mathbf{y}_n\) as ground truth for explanations. For each instance \(n\) and feature \(f\), the explainer (e.g., SHAP) gives a numeric importance \(e_{n,f}\).

    We collect all absolute importance values into the multiset
    \begin{equation}
    E = \bigl\{\lvert e_{n,f}\rvert : n=1,\dots,N,\;f=1,\dots,F\bigr\}.
    \end{equation}

    The global threshold \(T\) is set to the 90.00\textsuperscript{th} percentile of the distribution \(E\).
    \begin{equation}
        T = \mathrm{Percentile}(E, 90.00).
    \end{equation}
    This choice is grounded in our ablation study (see Appendix~\ref{appendix:ablation}), which demonstrates that the 90\textsuperscript{th} percentile yields the highest pattern granularity while optimizing computational time, avoiding the loss of informative signal observed at higher extremes.

    A feature \(f\) is \emph{important} for instance \(n\) if
    \begin{equation}
    \lvert e_{n,f}\rvert \;\ge\; T.
    \end{equation}
    Let \(F_{*}\) be the set of important features for instance \(n\).

    For each \(f\in F_{*}\), let \(x_{n,f}\) be the original feature value.  
    We compute the standard deviation \(\sigma_{f}\) of \(\{x_{n,f}: n=1,\dots,N\}\) on the training set. 

    We generate \(M\) perturbed samples jointly over all \(f\in F_{*}\):
    \begin{equation}
    x_{n,f}^{(m)} \sim \mathrm{Uniform}\bigl(x_{n,f} - \sigma_{f},\;x_{n,f} + \sigma_{f}\bigr).
    \end{equation}
    Each perturbed instance is fed to the model.

    We then find, for each feature \(f\), the smallest interval \((l_{f},h_{f}]\) such that every perturbed sample produces the same full prediction vector \(\mathbf{y}_n\). This enforces consistency across all output labels.
    In rare cases where no stable interval exists (i.e., all perturbations alter the prediction), the feature is excluded from the rule, as it indicates a highly sensitive decision boundary region where small changes cause class transitions.

    Finally, we define a rule for instance \(n\):
    \begin{equation}
    R_n: \bigwedge_{f\in F_{*}} \bigl(x_f \in (l_f, h_f]\bigr).
    \end{equation}

    Here, $R_n$ denotes the rule for instance $n$, $\bigwedge$ represents the logical conjunction over the set of important features $F_{*}$.
    For each important feature $f$, the value $x_f$ must fall within ($l_f$, $h_f$] to preserve the original prediction.

\subsection{Prototype Selection}
\label{sec:prototypes}
    In clinical ECG interpretation, practitioners require immediate explanations to support time-critical decisions. 
    However, counterfactual generation methods that rely on exhaustive search or complex optimizations at inference time can introduce latency incompatible with point-of-care requirements. 
    To address this challenge, we pre-compute representative prototypes for each diagnostic class to initialize counterfactual generation. 
    Our prototype selection process consists of four key steps: 
    \begin{enumerate}
        \item \textbf{Sample Filtering:} 
            Let $\mathcal{D}_{\text{train}} = \{(x_i, y_i)\}_{i=1}^N$ be the training set where $x_i \in \mathbb{R}^{T \times C}$ is a 12-lead ECG signal and $y_i \in \{0,1\}^L$ is its multi-label vector. For each class $l \in \{1, ..., L\}$, we define:

            \begin{equation}
            \mathcal{D}_l = \{x_i : (x_i, y_i) \in \mathcal{D}_{\text{train}}, y_{i,l} = 1, \sum_{j=1}^L y_{i,j} = 1\}
            \end{equation}

            Here, $\mathcal{D}_l$ denotes the filtered subset for class $l$, $(x_i, y_i)$ represents a sample-label pair from the training set $\mathcal{D}_{\text{train}}$, $y_{i,l}$ is the binary label for class $l$, and the summation condition $\sum_{j=1}^L y_{i,j} = 1$ enforces that the selected sample is single-label.
            This filters samples that belong exclusively to class $l$. 
            This design choice prioritizes generating clear, actionable counterfactuals that highlight minimal changes for a single diagnostic transition, aligning with how clinicians typically reason about differential diagnosis. 
            The comorbidity combinations are common (Table~\ref{tab:combination_distribution}), but treating each combination as a separate target class would exponentially increase prototype complexity and reduce interpretability. 
            Furthermore, filtering for single-class exclusivity implicitly removes ambiguous or potentially noisy recordings, ensuring that the resulting prototypes are stable and highly representative of their core pathology.

        \item \textbf{Distance-based Dimensionality Reduction:}
            In this step, we compute pairwise Dynamic Time Warping (DTW) distances between all training samples within each class, creating a distance matrix that captures morphological similarities between ECG signals.
            DTW was selected for its ability to accommodate natural heart rate variability, aligning corresponding waveform components even across patients with different cardiac rhythms.
            For two ECG signals $x_i, x_j \in \mathcal{D}_l$, the Dynamic Time Warping distance is:
            \begin{equation}
                \text{DTW}(x_i, x_j) = \min_{\pi \in \Pi} \sum_{(t,s) \in \pi} \|x_i(t) - x_j(s)\|_2
            \end{equation}
            where $\Pi$ is the set of all valid warping paths between the sequences.

            For class $l$, we construct the distance matrix $D_l \in \mathbb{R}^{|\mathcal{D}_l| \times |\mathcal{D}_l|}$ where:
            \begin{equation}
                D_l[i,j] = \text{DTW}(x_i, x_j) \quad \forall x_i, x_j \in \mathcal{D}_l
            \end{equation}

            We apply Multi-Dimensional Scaling to find a lower-dimensional embedding:
            $\min_{Z_l} \sum_{i,j} (||z_i - z_j||_2 - D_l[i,j])^2$
            where $Z_l = [z_1, ..., z_{|\mathcal{D}_l|}]^T \in \mathbb{R}^{|\mathcal{D}_l| \times d}$ and $d$ is chosen to maximize:
            \begin{equation}
                s(d) = \frac{1}{|\mathcal{D}_l|} \sum_{i=1}^{|\mathcal{D}_l|} \frac{a(i) - b(i)}{\max(a(i), b(i))}
            \end{equation}

            Where $z_i$ is the d-dimensional embedding of ECG sample $x_i$, and $Z_l$ is the matrix of all embeddings for class $l$, and $a(i)$ is the mean intra-cluster distance and $b(i)$ is the mean nearest-cluster distance.

        \item \textbf{Clustering in Multi-Dimensional Scaling (MDS) Space:}
            We partition $Z_l$ into $k$ clusters using k-means (the empirical selection of the hyperparameter $k$ via silhouette analysis is detailed in Section~\ref{sec:prototype_impl}):
                \begin{equation}
                    \{C_1^l, ..., C_k^l\} = \arg\min_{\mathcal{C}} \sum_{j=1}^k \sum_{z \in C_j} ||z - \mu_j||_2^2
                \end{equation}

            For each cluster $C_j^l$, we select the medoid as the prototype, the prototype $p_j^l$ for cluster $j$ of class $l$ is:
            \begin{equation}
                p_j^l = \arg\min_{x_i \in \mathcal{X}_{C_j^l}} \sum_{x_k \in \mathcal{X}_{C_j^l}} \text{DTW}(x_i, x_k)
            \end{equation}
            where $\mathcal{X}_{C_j^l} = \{x_i : z_i \in C_j^l\}$ maps points back to original ECG space.

            The final prototype set is $\mathcal{P} = \bigcup_{l=1}^L \bigcup_{j=1}^{k_l} \{p_j^l\}$. 

            \noindent By selecting medoids (actual training instances) rather than abstract centroids, we ensure that every explanation is anchored in a real, verifiable patient record. 
            This provides a traceable data origin essential for governance in safety-critical clinical systems, distinguishing our approach from generative methods that may synthesize plausible yet unverified signal patterns.

    \end{enumerate}

\subsection{From Prototypes to Sparse ECG Explanations} \label{sec:explanation} 
    Given a query ECG signal and its model prediction, our method generates counterfactual explanations through a two-stage process: prototype alignment followed by sparsification. 
    This approach ensures both temporal coherence and minimal perturbations.
    To ensure physiological plausibility, we implement a specialized segment cleaning procedure detailed in Section~\ref{sec:sparsity_opt}, which enforces a minimum length of $10$ samples to eliminate scattered or isolated modifications that lack clinical meaning. 
    Furthermore, by generating modifications across consecutive timestamps and anchoring the explanations in real-patient medoids, we ensure that the final insights are both stable and grounded in empirical cardiac morphologies rather than stochastic artifacts.

    \subsubsection{R-peak Based Temporal Alignment}\label{sec:r_aligement}
        ECG signals exhibit natural variability in heart rate and rhythm, even within the same diagnostic class. 
        Consequently, a direct element-wise comparison between a query and a prototype is often erroneous due to phase shifts or mismatches in the number of beats.
        To address this, we adopt a query-centric approach: since the explanation must interpret the specific patient's recording, we strictly maintain the temporal structure of the query signal. 
        To generate physiologically plausible counterfactuals that can be more easily analyzed by overlaying them, we align the temporal structure of selected prototypes to match the R-peaks of the query signal.  
        The alignment procedure warps the prototype's time axis to match the query's R-peak positions, ensuring that morphological comparisons focus on shape differences within corresponding cardiac cycles rather than rhythm discrepancies.
        We employ R-peak detection and piecewise-linear warping:

        \begin{enumerate}
            \item \textbf{R-peak Detection}: R-peaks define the cardiac cycle boundaries and are detected in both the query signal and the selected prototype using either WFDB's XQRS algorithm or NeuroKit2's peak detection, depending on signal quality. 

            \item \textbf{Peak Count Normalization}: If the prototype and query have different numbers of detected beats:
            \begin{itemize}
                \item When the prototype has fewer beats, we pad the regions preceding the first and following the last matching R-peak with zeros to match the query duration. 
                In the subsequent alignment step, we {only warp the beats that correspond to those in the query signal}, leaving the padded boundary regions flat.
                Although this introduces local non-physiological artifacts, it ensures that the complete diagnostic context of the query is preserved. 
                Crucially, these artifacts are typically discarded during the sparsity optimization (Section~\ref{sec:sparsity_opt}), as they do not contribute to the target classification.
                \item When the prototype has more beats, we truncate the preceding cardiac cycles to match the query's beat count. 
                However, as the piecewise-linear warping is mathematically defined strictly between corresponding R-peaks, the signal regions outside this interval, specifically before the first and after the last matched peak—are {set to zero}.
                These resulting flat boundaries are unavoidable in peak-to-peak alignment but are subsequently discarded by the sparsity optimization, ensuring the final explanation remains physiologically plausible.
            \end{itemize}
            \item \textbf{Piecewise-Linear Warping}: For each inter-peak segment and each ECG lead independently, we apply linear interpolation to map the prototype segment to the corresponding query segment. 
            This preserves the morphology of individual cardiac cycles while adapting to the query's specific rhythm.
        \end{enumerate}

        The warping function for segment $k$ between R-peaks is defined as:
        \begin{equation}
            \mathbf{x}_{\text{aligned}}[t] = \mathbf{x}_{\text{proto}}\left( s_k + \frac{s_{k+1} - s_k}{t_{k+1} - t_k} (t - t_k) \right)
        \end{equation}
        where $t_k, t_{k+1}$ are consecutive R-peak positions in the query, and $s_k, s_{k+1}$ are the corresponding peaks in the prototype.
        Since the mapped time coordinate is continuous, the value of $\mathbf{x}_{\text{proto}}$ is evaluated via linear interpolation.
        This linear scaling effectively stretches or compresses the prototype segment to perfectly match the duration of the query's cardiac cycle.

    \subsubsection{Sparsity Optimization}\label{sec:sparsity_opt}

        The aligned prototype serves as an initial counterfactual, but since it originates from a different patient recording, it inherently differs from the query at nearly every time point despite temporal alignment. To enhance interpretability and identify the minimal changes required for class transition, we apply sparsity optimization:

        \begin{enumerate}
            \item \textbf{Importance scoring}: We compute the absolute difference between the aligned prototype and query at each time-channel point. 
            Crucially, R-peak regions receive doubled importance scores to prioritize central cardiac features over peripheral signal components.
            Consequently, the flat boundary artifacts introduced by alignment (Section~\ref{sec:r_aligement}) typically yield lower weighted scores compared to high-magnitude diagnostic discrepancies (e.g., ST-segment deviations), causing them to be naturally deprioritized during selection.

            \item \textbf{Iterative sparsification}: Starting with a low keep-ratio (e.g., 10\%), we iteratively determine which signal segments to modify:
            \begin{itemize}
                \item Create a binary mask based on importance score percentiles
                \item Replace only the masked regions of the query with the corresponding prototype values
                \item Verify that the model's prediction matches the target class
                \item If successful, attempt a sparser solution; otherwise, increase the keep-ratio
            \end{itemize}

            \item \textbf{Segment Cleaning}: We enforce minimum segment lengths (10 samples) to avoid scattered, physiologically implausible modifications. Small isolated changes are removed from the modification mask.
        \end{enumerate}

        The optimization objective is formulated as:
        \begin{equation}
            \min_{\mathbf{m}} ||\mathbf{m}||_0 \quad \text{s.t.} \quad \mathcal{M}(\mathbf{x}_q \odot (1-\mathbf{m}) + \mathbf{x}_p \odot \mathbf{m}) = y_{\text{target}}
        \end{equation}
        where $\mathbf{m}$ is the binary modification mask, $\mathbf{x}_q$ is the query, $\mathbf{x}_p$ is the aligned prototype, $\odot$ denotes element-wise multiplication, and $\mathcal{M}$ is the model.

    \subsubsection{Query Processing Workflow}
        \begin{figure}[h!]
            \centering
            \includegraphics[width=\textwidth]{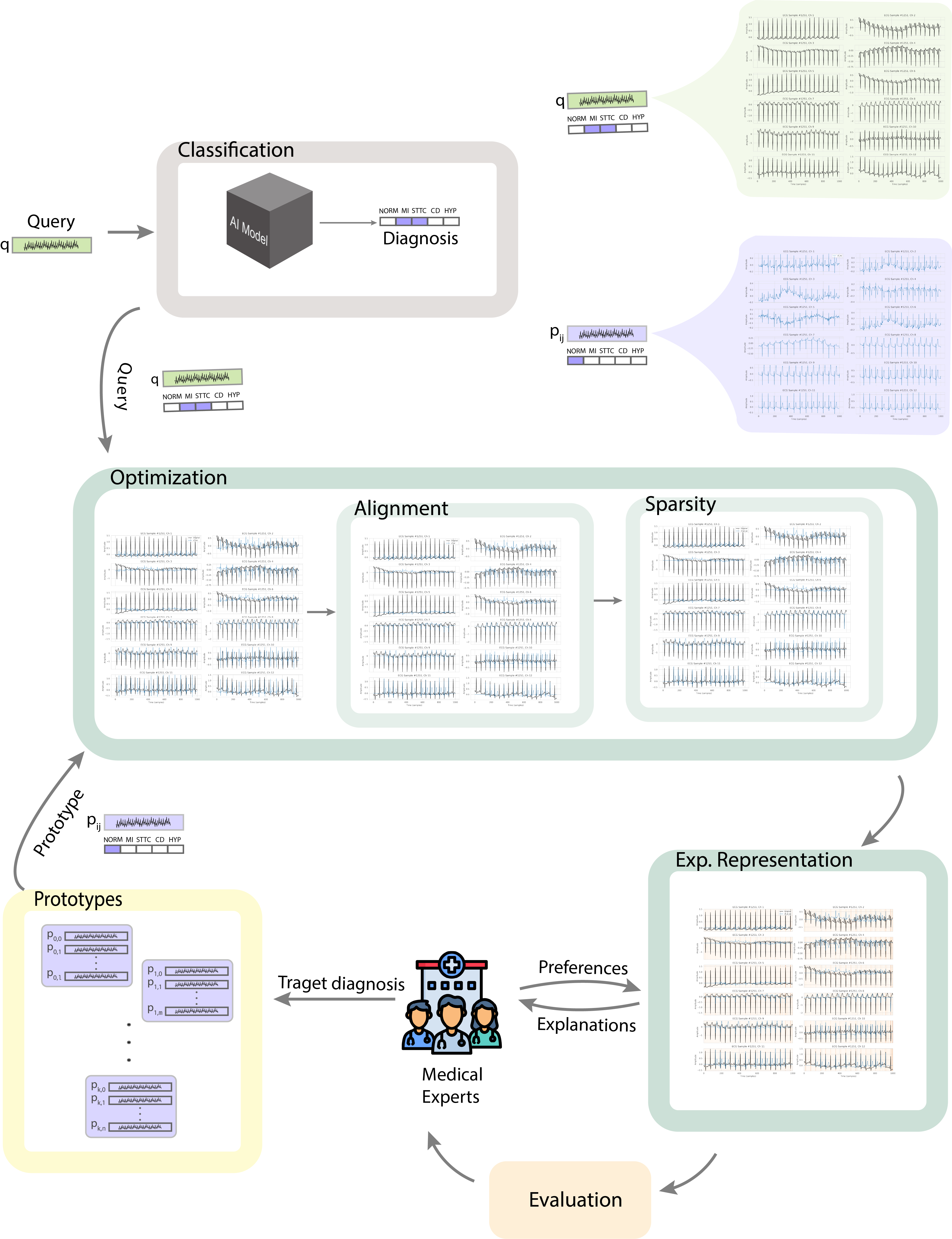}
            \caption{ECG explanation generation pipeline: from query ECG classification through prototype retrieval, R-peak alignment, and sparsity optimization to final counterfactual visualization.}
            \label{fig:flow}  
        \end{figure}
        The complete query processing workflow, illustrated in Figure~\ref{fig:flow}, integrates the prototype selection methodology described in Section~\ref{sec:prototypes} with the alignment and sparsification techniques detailed above. This end-to-end pipeline transforms a query ECG and model prediction into an interpretable counterfactual explanation. When a new query ECG requires explanation:
        \begin{enumerate}
            \item Take the ECG signal, predicted classes by the AI model, and the target counterfactual class as input. 
            If the target class is not provided, identify appropriate counterfactual classes based on the model's prediction confidence. 
            The system prioritizes transitions to neighboring classes in the diagnostic hierarchy (e.g., from pathological to normal) to keep clinical meaningfulness.
            \item Retrieve the pre-computed prototypes for the target class, which were extracted during the offline MDS-based clustering phase. 
            Select the prototype with minimal DTW distance to the query. 
            As discussed in Section~\ref{sec:bg}, we prioritize existing patient medoids over synthetic averages (barycenters) to ensure that the baseline explanation remains physiologically authentic and retains all high-frequency clinical details.
            \item Apply the two-stage optimization process: (i) R-peak alignment to temporally warp the prototype, adapting its rhythm to match the query's cardiac cycle, and (ii) sparsity optimization to identify the minimal set of modifications required for class transition while preserving physiological plausibility.
            \item Return the sparse counterfactual along with visualizations highlighting the specific ECG regions that drive the diagnostic change, providing suitable visualizations to clinicians to focus on the most relevant waveform alterations.
        \end{enumerate}

        This approach generates explanations in under a second per 10-second ECG recording (1000 samples at 100Hz) by using pre-computed prototypes. 
        The two-stage process, temporal alignment followed by sparse modifications, produces counterfactuals that respect ECG physiology. 
        By preparing prototypes offline and running only lightweight optimizations at query time, we achieve both speed and quality.
        This design specifically targets the low-latency requirements of real-time monitoring systems, achieving a balance between physiological fidelity and the rapid response times necessary for point-of-care deployment.

\subsection{Objective Evaluation}

    To comprehensively assess the quality of generated counterfactual explanations, we developed five key evaluation metrics for 12-lead ECG explanations, which are commonly used in the literature for tabular counterfactuals (\cite{bayrak2024evaluation}), which we adapted for the specific characteristics of multivariate timeseries ECG signals. Unlike tabular data, where features are independent, ECG signals exhibit temporal dependencies, morphological constraints, and channel correlations that require specialized evaluation approaches.

    \subsubsection{Validity}
        Validity is the fundamental requirement for any counterfactual explanation, measuring whether it successfully changes the model's prediction to the desired target class. For ECG signals, we must ensure that the counterfactual not only crosses the decision boundary but also maintains physiological plausibility. Given a query ECG signal $\mathbf{x} \in \mathbb{R}^{T \times C}$ (where $T=1000$ samples and $C=12$ leads) with model prediction $\mathcal{M}(\mathbf{x}) = y$, and its counterfactual $\mathbf{x}'$, validity is defined as:

        \begin{equation}
        \text{validity}(\mathbf{x}, \mathbf{x}') = \begin{cases}
        1 & \text{if } \mathcal{M}(\mathbf{x}') = y_{\text{target}} \land \mathcal{M}(\mathbf{x}') \neq \mathcal{M}(\mathbf{x}) \\
        0 & \text{otherwise}
        \end{cases}
        \end{equation}

        For multi-label ECG classification, we extend this to ensure all class predictions change appropriately:
        \begin{equation}
        \text{validity}_{\text{multi}} = \prod_{i=1}^{L} \mathbb{1}[\mathcal{M}_i(\mathbf{x}') = y_{\text{target},i}]
        \end{equation}

        where $L$ is the number of diagnostic labels, $\prod$ denotes the product over all classes (equivalent to a logical AND), $\mathbb{1}[\cdot]$ is the indicator function returning 1 if the condition holds and 0 otherwise, $\mathcal{M}_i(\mathbf{x}')$ represents the model's binary prediction for class $i$ given counterfactual $\mathbf{x}'$, and $y_{\text{target},i}$ is the desired label for that class.

    \subsubsection{Sparsity}
        Sparsity quantifies the locality and efficiency of modifications, crucial for interpretability in clinical settings (\cite{bayrak2023pertcf}). Sparse counterfactuals highlight specific ECG features (e.g., ST-segment changes, QRS morphology) responsible for diagnostic decisions. We evaluate sparsity through multiple complementary metrics:

        \begin{itemize} 

            \item \textbf{Sparsity Ratio} measures the fraction of modified samples across all channels:
            \begin{equation}
            \text{sparsity\_ratio} = \frac{1}{T \times C} \sum_{t=1}^{T} \sum_{c=1}^{C} \mathbb{1}[|\mathbf{x}'_{t,c} - \mathbf{x}_{t,c}| > \tau]
            \end{equation}
            where $T$ and $C$ denote the number of time points and leads respectively, 
            $\mathbf{x}'_{t,c}$ and $\mathbf{x}_{t,c}$ are the signal values at time $t$ and lead $c$ for the counterfactual and query, and $\tau = 0.01 \cdot \sigma_{\mathbf{x}}$ is the noise tolerance threshold derived from the training data standard deviation $\sigma_{\mathbf{x}}$.

            \item \textbf{L$_p$ Norms} provide alternative sparsity measures:
            \begin{align}
            \text{sparsity}_{L_0} &= \|\mathbf{x}' - \mathbf{x}\|_0 = \sum_{t,c} \mathbb{1}[|\mathbf{x}'_{t,c} - \mathbf{x}_{t,c}| > 0] \\
            \text{sparsity}_{L_1} &= \|\mathbf{x}' - \mathbf{x}\|_1 = \sum_{t,c} |\mathbf{x}'_{t,c} - \mathbf{x}_{t,c}| \\
            \text{sparsity}_{L_2} &= \|\mathbf{x}' - \mathbf{x}\|_2 = \sqrt{\sum_{t,c} (\mathbf{x}'_{t,c} - \mathbf{x}_{t,c})^2}
            \end{align}
        \end{itemize}

    \subsubsection{Stability}
        Stability metrics assess the robustness of counterfactual explanations against minor perturbations (\cite{darias2024empirical}), crucial for clinical reliability where measurement noise and physiological variations are inherent.

        \begin{itemize}
            \item \textbf{Noise Stability} tests robustness under Gaussian perturbations:
            \begin{equation}
            \text{stability\_score} = \frac{1}{|\mathcal{N}| \cdot N_{\text{trials}}} \sum_{i=1}^{N_{\text{trials}}} \sum_{\sigma_{\epsilon} \in \mathcal{N}} \mathbb{1}[\mathcal{M}(\mathbf{x}' + \epsilon_i) = \mathcal{M}(\mathbf{x}')]
            \end{equation}
            where $\epsilon_i \sim \mathcal{N}(0, \sigma_{\epsilon}^2\mathbf{I})$ and $\mathcal{N} = \{0.01, 0.02, 0.05\} \cdot \text{std}(\mathbf{x}')$.

            \item \textbf{Temporal Stability} evaluates invariance to small time shifts, important for ECG signals where slight phase variations are common:
            \begin{equation}
            \text{temporal\_stability} = \frac{1}{1 + \overline{\text{DTW}}_{\tau}}
            \end{equation}
            where:
            \begin{equation}
            \overline{\text{DTW}}_{\tau} = \frac{1}{|\mathcal{T}|} \sum_{\tau \in \mathcal{T}} \frac{\text{DTW}(\mathbf{x}', \text{shift}(\mathbf{x}', \tau))}{\text{DTW}_{\text{norm}}}
            \end{equation}
            with $\mathcal{T} = \{-2, -1, 1, 2\}$ samples and $\text{DTW}_{\text{norm}} = \sqrt{T \cdot C}$ for normalization.

        \end{itemize}

    \subsubsection{Decision Margin}
        The decision margin quantifies how confidently the counterfactual achieves the target classification, providing a buffer against classification boundary fluctuations:
        \begin{equation}
        \text{decision\_margin} = p_{\mathcal{M}(\mathbf{x}')}(\mathbf{x}') - t_{\mathcal{M}(\mathbf{x}')}
        \end{equation}
        where $p_{\mathcal{M}(\mathbf{x}')}(\mathbf{x}')$ is the predicted probability and $t_{\mathcal{M}(\mathbf{x}')}$ is the class-specific threshold (which may differ from 0.5 in multi-label settings).

    These metrics collectively measure that the generated counterfactual explanations meet the requirements of clinical ECG interpretation, such as they must be valid, sparse enough to be interpretable, stable under realistic variations, and confident in their predictions. These evaluation metrics allow systematic comparison and selection of high-quality counterfactual explanations for 12-lead ECG classification models.

    \section{Experiments}\label{sec:experiments}
    \label{sec:exp}

    \subsection{Data}

    The PTB-XL dataset contains 21\,799 clinical 12-lead ECG recordings sampled at 100 Hz as mentioned in~\cite{wagner_2022_ptbxl}. 
    Each recording lasts 10s and includes patient metadata, signal-quality annotations, and diagnostic labels. 
    PTB-XL is available via PhysioNet~(\cite{goldberger_2000_physionet,wagner_2022_ptbxl}) and was first described in Scientific Data by~\cite{wagner_2020_ptbxl}. 

    Table~\ref{tab:class_distribution} shows the distribution of the five main diagnostic classes. 
    Normal ECGs (NORM) are most common, followed by myocardial infarction (MI), ST/T changes (STTC), conduction disturbances (CD) and hypertrophy (HYP).     
    Crucially, these diagnostic categories primarily reflect morphological and structural abnormalities (e.g., changes in wave amplitude, segment deviation, or impulse conduction time) rather than chaotic rhythm disorders. 
    Consequently, the vast majority of recordings in this subset retain identifiable cardiac cycles with stable R-peaks, supporting the applicability of the peak-based alignment strategy described in Section~\ref{sec:method}.
    \begin{table}[ht]
      \centering
      \caption{Distribution of primary diagnostic classes (n=21\,799).}
      \label{tab:class_distribution}
      \begin{tabular}{lrr}
        \toprule
        Class & Percentage (\%) & Count \\
        \midrule
        NORM & 43.64 & 9\,514 \\
        MI   & 25.09 & 5\,469 \\
        STTC & 24.01 & 5\,235 \\
        CD   & 22.47 & 4\,898 \\
        HYP  & 12.15 & 2\,649 \\
        \bottomrule
      \end{tabular}
      \vspace{0.5em}

      {\footnotesize
        \textbf{Abbreviations:}
        NORM – Normal ECG;
        MI – Myocardial Infarction;
        STTC – ST/T Change;
        CD – Conduction Disturbance;
        HYP – Hypertrophy.
      }
    \end{table}

    Table~\ref{tab:combination_distribution} lists the most frequent diagnostic label combinations of at least 2 labels in PTB-XL, which exceeds 1.0\%. 
    The most common pairing is MI with CD (8.23\%), followed by STTC with HYP (6.92\%) and MI with STTC (6.14\%), which are most prevalent overlapping pathologies. 
    \begin{table}[ht]
      \centering
      \caption{Diagnostic label combinations exceeding 1\,\% in PTB-XL.}
      \label{tab:combination_distribution}
      \begin{tabular}{lrr}
        \toprule
        Combination        & Count & Percentage (\%) \\
        \midrule
        MI, CD             & 1794 & 8.23 \\
        STTC, HYP          & 1509 & 6.92 \\
        MI, STTC           & 1339 & 6.14 \\
        STTC, CD           & 1066 & 4.89 \\
        MI, HYP            &  818 & 3.75 \\
        HYP, CD            &  787 & 3.61 \\
        MI, STTC, HYP      &  517 & 2.37 \\
        NORM, CD           &  415 & 1.90 \\
        MI, STTC, CD       &  379 & 1.74 \\
        STTC, HYP, CD      &  367 & 1.68 \\
        MI, HYP, CD        &  274 & 1.26 \\
        \bottomrule
      \end{tabular}
      \vspace{0.5em}

      {\footnotesize
      \textbf{Abbreviations:} 
      NORM – Normal ECG; 
      MI – Myocardial Infarction; 
      STTC – ST/T Change; 
      CD – Conduction Disturbance; 
      HYP – Hypertrophy.
      }
    \end{table}

    \subsection{Data Preprocessing} \label{sec:data_preprocessing}
    The authors of PTB-XL provided stratified sample folds, which we adopted without modification. 
    In our case, folds 1–7 was held out for training and fold 9 for validation.
    For fold 10 we calculated SHAP values, which then were used do create rules in~\emph{PHAR} method and further clustered to obtain medoids (which acted as counterfactuals). 
    Additionally fold 7 was used for calculating statistics for~\emph{PHAR} method and 
    fold 9 was used as a background for calculating SHAP values. 
    Fold 8, not used in any other place, was held out as an independent test set, providing unseen query samples to evaluate the generalization capability of the entire explanation pipeline.
    This split yielded 15 245 training samples (69.9\%), 2 183 validation samples (10.0\%), 2 198 samples to create rules and clustering them (10.1\%) and 2 173 query samples (10.0\%).

    All ECG signals were normalized using statistics computed on the training set. 
    We found the mean of \(X_{\mathrm{train}}\) to be \(\mu = -0.0008313\) and its standard deviation \(\sigma = 0.2357998\). 
    Each signal \(x\) was then scaled as \(x_{\mathrm{norm}} = \frac{x - \mu}{\sigma + 10^{-7}}\).

    Diagnostic labels were converted to multi-hot encodings for deep‐network training. 
    We applied a \textit{multi-label binarizer} to map each list of labels into a fixed five‐dimensional binary vector, where each position corresponds respectively to Normal ECG, Myocardial Infarction, ST/T Change, Conduction Disturbance and Hypertrophy. 
    These multi-hot vectors served as the target outputs during model training.

    \subsection{Model}
\begin{figure}[!t]  
            \centering
            \includegraphics[width=\textwidth]{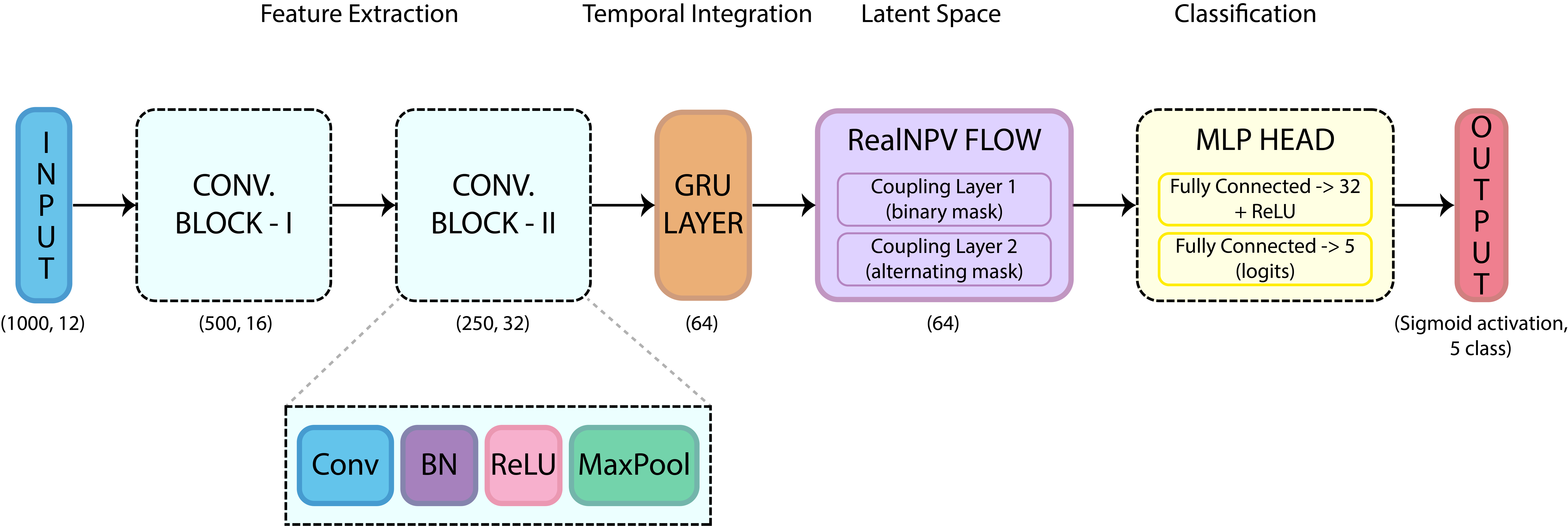}
            \caption{12-lead ECG multiclass classification architecture: raw signals (1000 time points \(\times\) 12 leads) are processed through convolutional feature extraction (\textit{blue}), batch normalization (\textit{purple}), ReLU activations (\textit{pink}), and max-pooling (\textit{green}); then temporally modeled by a GRU (\textit{orange}), transformed in latent space via RealNVP flow (\textit{lavender}), and classified with an MLP head (\textit{yellow}).}
            \label{fig:model_architecture}  
        \end{figure}

    In Figure~\ref{fig:model_architecture}, we present the architecture of our 12-lead ECG multiclass classification model. 
    The network accepts raw ECG inputs of shape \(1000~time~points~\times~12~leads\). 
    First, two convolutional blocks extract local waveform features; each block combines convolutional layers (\textit{blue}), batch normalization (\textit{purple}), ReLU activations (\textit{pink}), and max-pooling (\textit{green}). 
    The encoded features are then fed into a GRU layer (\textit{orange}) for temporal integration. 
    Next, a RealNVP normalizing flow (\textit{lavender}) transforms the GRU output into a latent space. 
    Finally, an MLP head (\textit{yellow}), consisting of two fully connected layers, maps the latent representation to class logits and produces probability scores via sigmoid activation.

    \subsubsection{Model architecture}
    The network comprises convolutional, recurrent, flow-based and feed-forward components. Specifically:
    \begin{itemize}
      \item \textbf{Input:} Raw ECG signal of shape $(1000,12)$.
      \item \textbf{Conv Block 1:} Conv1D with 16 filters (kernel size 5, padding 2), followed by batch normalization, ReLU activation and max-pooling, yielding output of shape $(500,16)$.
      \item \textbf{Conv Block 2:} Conv1D with 32 filters (kernel size 5, padding 2), followed by batch normalization, ReLU activation and max-pooling, yielding output of shape $(250,32)$.
      \item \textbf{GRU Layer:} Processes the sequence of length 250 with 32-dimensional inputs to produce a fixed embedding of size $H$.
      \item \textbf{RealNVP Flow:} Two coupling layers with alternating binary masks; each layer uses MLPs to scale and translate one half of the $H$-dimensional vector, producing a latent vector of the same size.
      \item \textbf{MLP Head:} Fully connected layer to 32 units with ReLU, then a second layer to $C$ logits for multi-label classification.
    \end{itemize}
    In our model, the embedding dimension is \(H = 64\) and the number of classes is \(C = 5\).

    \textit{Convolutional blocks} extract local waveform features and reduce the temporal dimension by scanning the signal with small filters, stabilizing activations via batch normalization, injecting non-linearity with ReLU, and halving the time length through max-pooling.
    \textit{GRU layer} integrates these features over time, retaining important patterns and discarding noise.
    \textit{RealNVP flow} warps the embedding distribution into a flexible latent space by alternately scaling and shifting parts of the vector under learned transformations. 
    While our ablation study indicated comparable classification performance without this block, we retained it to ensure robust modeling of complex probability densities, following the literature discussed in Section~\ref{sec:DL_for_ECG}.
    \textit{MLP head} maps the latent vector to class logits, which are converted to independent probabilities by sigmoid at inference.

    \subsection{Training}\label{subsec:training}
    We trained the network with Adam (LR$=10^{-4}$) for up to 1000 epochs, applying early stopping after 25 non-improving epochs and gradient norm clipping to prevent exploding updates. 
    This setup ensured fast convergence while avoiding overfitting, as evidenced by the loss curves in Figure~\ref{fig:training-history}. 
    An error-pattern analysis (Table~\ref{tab:misclass}) highlights the most frequent confusions and guides future refinements. 
    For full experimental details and additional plots, see Appendix~\ref{appendix:training}.

    \subsection{Threshold selection}\label{subsec:threshold_selection}
    To balance precision and recall across imbalanced classes, we determined optimal decision thresholds on the validation set by sweeping probabilities to maximize each class’s F1 score. 
    Adopting these tailored cut-offs improved the model’s overlap measures (notably Jaccard and F1) on test data while only slightly reducing exact‐match accuracy (see Figure~\ref{fig:roc-per-class}). 
    Detailed methodology and threshold values are available in Appendix~\ref{appendix:threshold_selection}. 

    \subsection{SHAP value computation}
    We used the SHAP library\footnote{\url{https://shap.readthedocs.io/en/latest/}} (version 0.47.0) and its \textit{GradientExplainer} to attribute model predictions to input features. 
    \textit{GradientExplainer} combines model gradients with a background distribution to efficiently approximate Shapley values. 
    A representative subset (fold 8) of the data served as the background distribution. 
    For each batch of ECG segments (shape $(B,1000,12)$), the explainer produced raw SHAP values of shape $(B,1000,12,5)$ - one value per time step, channel, and output class - which were then transposed to $(B,5,1000,12)$ for consistency with our downstream analysis. 
    To handle large test sets efficiently, SHAP values were written incrementally into an HDF5 dataset, allowing interrupted runs to resume without loss. 

    \subsubsection{Rule extraction from SHAP explanations}
    The following steps follow the \emph{PHAR}~\citep{mozolewski2025phar} methodology for converting numeric attributions into interval rules. 

    To ensure that generated rules reflect the model’s actual decisions, we first convert the model’s logits into binary predictions using the previously determined per‐class thresholds. 
    We then iterate over each class that the model predicts as present, slicing out its corresponding SHAP explanation matrix of shape $(N, T, M)$ from the full $(N, C, T, M)$ array. 
    Rules are only extracted for these positively predicted classes.

    Important features - the ones which will be included in the rules for given data instance - are chosen by applying a single global cutoff at the specified percentile of all absolute SHAP values. 
    In our implementation we used \(percentile=90.0\). 

    For each selected time‐channel coordinate, we generate 1000 perturbed samples by adding Gaussian noise scaled by \(\sigma=1.0\). 
    We enforce that the entire multi‐label output remains unchanged under these perturbations. 
    Enforcing it means that when we derive a rule for a particular instance and class, we only accept intervals for which every perturbed version of that instance yields the exact same full set of predicted labels. 
    In other words, the rule for one class cannot flip any other class’s prediction on that example - so if an instance is assigned to multiple classes, each class’s rule will remain compatible and never contradict the others. 

    The maximal range of values that preserve consistency defines a simple interval rule (e.g.\ ``time\_t\_chan\_m \(\in\)(a,b]''). 
    Finally, we evaluate each rule’s coverage (fraction of samples satisfying the interval) and confidence (fraction of those whose full prediction vector matches the original).
    The result is a compact set of interval rules, each annotated with its empirical support and reliability.

\subsection{Prototype Selection Implementation}

    We implemented the prototype selection pipeline on the \emph{2,198} samples from an unused fold of the data. The process consisted of the following steps:

    \subsubsection{Sample Filtering}
        From the initial dataset, we retained only samples that met two criteria: 
        (i) correct classification by our trained model, and 
        (ii) single-label assignment. 
        This filtering resulted in \emph{947} samples distributed as: 
        NORM (\emph{377/380}, \emph{99.21\%}), 
        MI (\emph{160/317}, \emph{50.47\%}), 
        STTC (\emph{231/537}, \emph{43.02\%}), 
        CD (\emph{156/330}, \emph{47.27\%}), and 
        HYP (\emph{23/845}, \emph{2.72\%}). 
        Overall retention was \emph{947/2,198} (\emph{43.08\%}). 
        (Note: the sum of the per-class denominators is \emph{2,409}, but only \emph{2,198} unique samples existed due to multi-class overlaps).
        The filtering ensures prototype quality by excluding ambiguous or potentially mislabeled cases.

    \subsubsection{DTW Distance Computation}
        For each diagnostic class, we computed pairwise DTW distances between all samples. 
        We used the \emph{similarity-ts}\footnote{Available at \url{https://github.com/alejandrofdez-us/similarity-ts}} package implementation with 
        the \emph{DTW} metric and a stride of \emph{1}
        to reduce computational complexity while maintaining accuracy. 
        The DTW distance matrices captured morphological similarities between ECG waveforms within each class. 
        Distance computation took approximately \emph{89} hours for all classes on a system with \emph{64} CPU cores, however, the code was not specifically parallelized, and the libraries used (per their documentation) do not utilize multithreading. 

    \subsubsection{MDS Projection and Clustering}\label{sec:prototype_impl}
        We applied Multi-Dimensional Scaling to each class's DTW distance matrix. The optimal number of MDS components was determined using silhouette analysis:
        \begin{itemize}
            \item NORM: \emph{6} components (silhouette score: 0.\emph{44})
            \item MI: \emph{7} components (silhouette score: 0.\emph{32})
            \item STTC: \emph{6} components (silhouette score: 0.\emph{36})
            \item CD: \emph{5} components (silhouette score: 0.\emph{0.32})
            \item HYP: \emph{3} components (silhouette score: 0.\emph{28})
        \end{itemize}

\begin{figure}[h]
          \centering
          \includegraphics[width=0.8\linewidth]{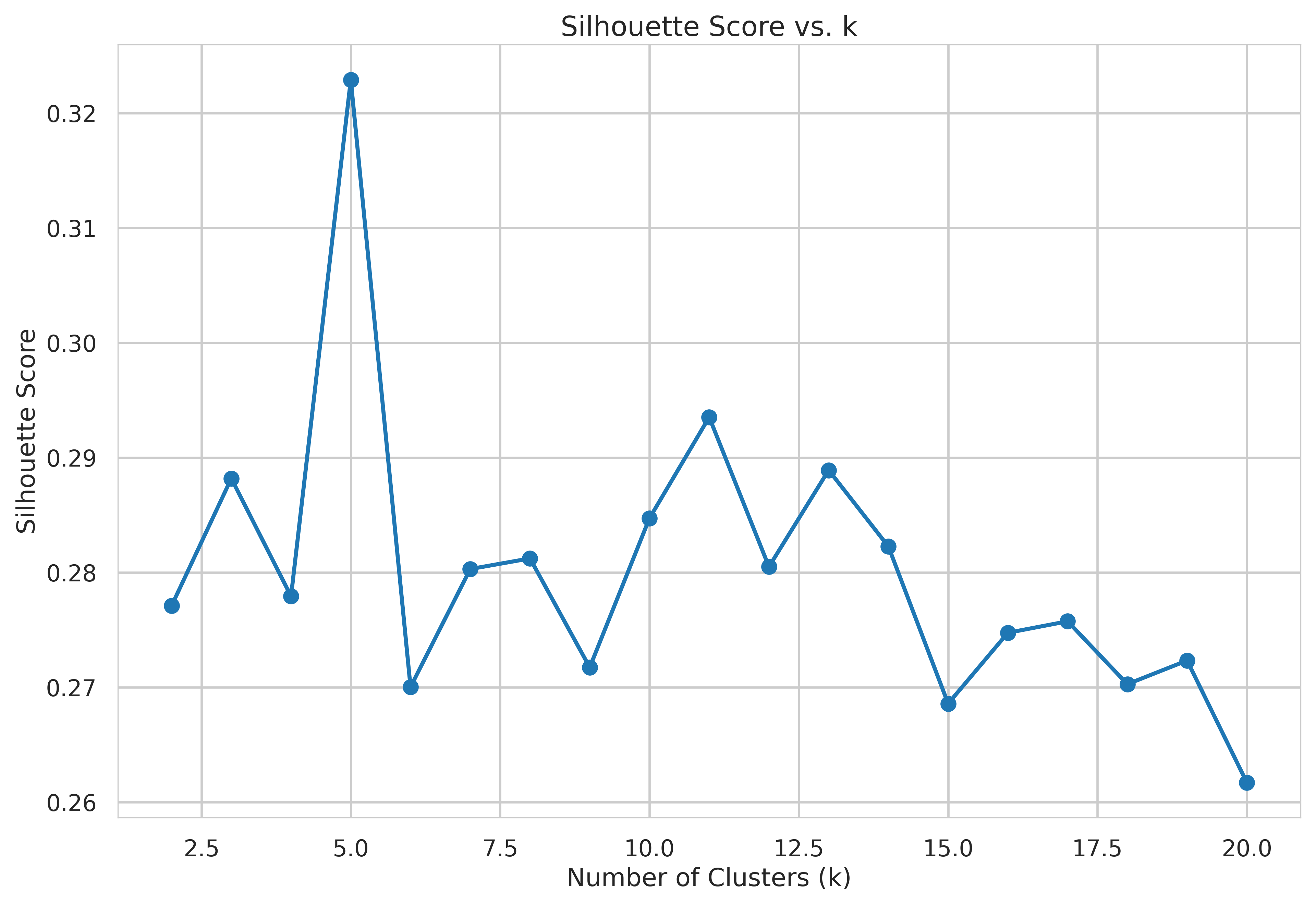}
          \caption{Silhouette analysis for optimal cluster selection in the Conduction Disturbance (CD) class.}
          \label{fig:cl0_silhouette}
        \end{figure}

        For k-means clustering in MDS space, we evaluated $k \in \{2, 3, \emph{...}, 10\}$ using combined metrics of silhouette coefficient and the optimal cluster counts were: NORM ($k=$\emph{6}), MI ($k=$\emph{7}), STTC ($k=$\emph{6}), CD ($k=$\emph{5}), and HYP ($k=$\emph{3}). 
        Figure~\ref{fig:cl0_silhouette} illustrates this optimization process for the CD class as a representative case, where the peak at $k=5$ indicates the optimal number of clusters. 
        The resulting clustering, visualized in Figure~\ref{fig:cl0_scatter}, shows separation of CD samples into four distinct morphological groups

    \subsubsection{Medoid Selection}

\begin{figure}[h]
          \centering
          \includegraphics[width=0.8\linewidth]{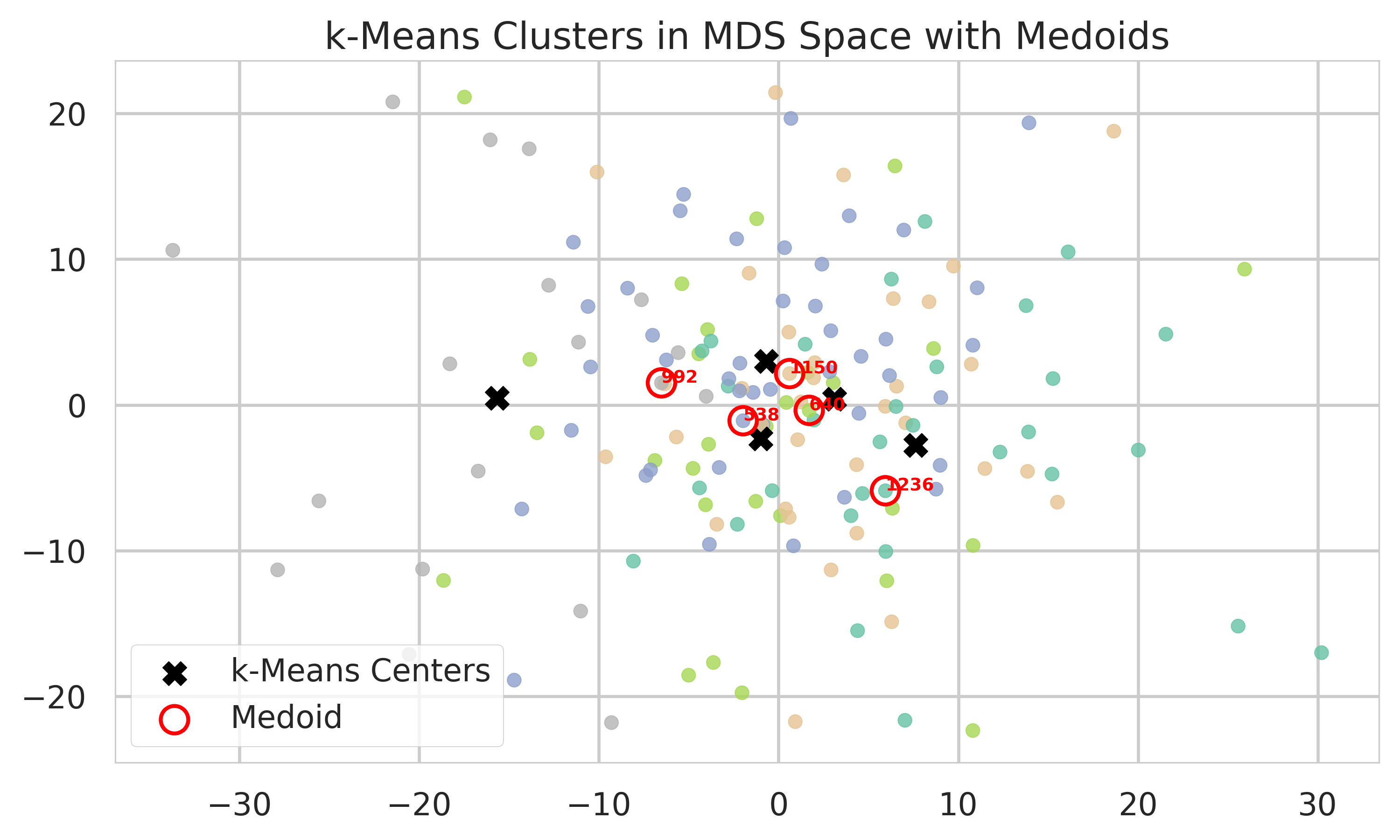}
          \caption{k-Means clustering results in MDS space for Conduction Disturbance (CD) class.}
          \label{fig:cl0_scatter}
        \end{figure}

        From each cluster, we identified the medoid as the sample with the minimum average DTW distance to all other cluster members. As shown in Figure~\ref{fig:cl0_scatter} for the CD class, medoids (marked with red circles) represent the most central examples within each cluster, ensuring representativeness. This resulted in 27 total prototypes across all classes, stored as a prototype database for counterfactual generation close to real-time.

    \section{Results and Discussion}\label{sec:discussion}
    \label{sec:disc}  

\subsection{Experimental Results}
    \label{subsec:res}
    This section presents a comprehensive evaluation of our prototype-driven counterfactual generation framework across three distinct approaches: \textit{Original} (prototype without optimization), \textit{Sparse} (prototype with sparsity optimization), and \textit{Aligned Sparse} (prototype with both R-peak alignment and sparsity optimization). Our evaluation employs five key metrics specifically adapted for multivariate time-series ECG data: \textit{validity}, \textit{sparsity} ($L_0, L_1, L_2$), \textit{stability} (noise and temporal), and \textit{decision margin}.

    \subsubsection{Overall Performance Analysis}

        \begin{table} [h!]
        \caption{Orginal vs counterfactuals: summary statistics of validity, sparsity, stability, and decision margin.}
        \label{tab:summary_overall}
        \begin{tabular}{lrrr}
        \toprule
        \textbf{Metric} & \textbf{Original} & \textbf{Sparse} & \textbf{Aligned Sparse }\\
        \midrule
        \textbf{Validity$_{multi}$} & \textbf{1.0000} & \textbf{1.0000} & 0.8126 \\
        \midrule
        \multicolumn{4}{l}{\textbf{Sparsity}} \\
        \quad Sparsity ratio & 0.9837 & 0.7957 & \textbf{0.7800} \\
        \quad $L_0$ sparsity & 0.9837 & 0.7957 & \textbf{0.7800} \\
        \quad $L_1$ sparsity & 0.2868 & 0.2700 & \textbf{0.2018} \\
        \quad $L_2$ sparsity & 0.2512 & 0.2442 & \textbf{0.1749} \\
        \midrule
        \multicolumn{4}{l}{\textbf{Stability}} \\
        \quad Noise stability & \textbf{1.0000} & \textbf{1.0000} & 0.9948 \\
        \quad Temporal stability & 0.5304 & 0.5097 & \textbf{0.7593} \\
        \midrule
        \textbf{Margin} & \textbf{0.7800} & 0.7750 & 0.6933 \\
        \bottomrule
        \end{tabular}
        \footnotesize Bold values indicate best performance per metric. For sparsity metrics, lower values are better (sparser counterfactuals). For all other metrics, higher values are better.
        \end{table}

        Table~\ref{tab:summary_overall} presents the aggregate performance across all 963 test samples, revealing distinct trade-offs between the three approaches. 
        The validity metric, fundamental to any counterfactual explanation, shows perfect performance (1.0) for both the Original and Sparse methods, indicating that the generated counterfactuals successfully change the model's prediction to the desired target class. However, the Aligned Sparse method shows a notable decrease in validity to 0.8126, representing an 18.74\% drop. 
        This represents an inherent trade-off between improved sparsity and temporal stability, at the cost of reduced validity. Importantly, our framework provides all three variants (Original, Sparse, and Aligned Sparse), allowing users to select the one that best suits their priorities. Perfect validity when explanation correctness is critical, or improved interpretability when physiological plausibility is preferred. For HYP specifically, the severe validity drop (13.2\%) suggests the Sparse variant without alignment may be more appropriate until alignment techniques better accommodate this class.

        \textit{Sparsity} performance reveals the effectiveness of our optimization approach. The sparsity ratio improves significantly from 0.9837 (Original) to 0.7800 (Aligned Sparse), indicating that our method modifies only 78\% of the original signal rather than nearly the entire waveform. This 20.7\% improvement in sparsity directly translates to enhanced interpretability, as clinicians can focus on specific ECG segments rather than examining wholesale signal replacements. The $L_1$ and $L_2$ sparsity metrics show consistent improvements, with $L_2$ sparsity decreasing from 0.2512 to 0.1749, demonstrating that modifications are not only fewer but also smaller in magnitude.

        \textit{Stability} analysis reveals complementary strengths across methods. Noise stability remains consistently high ($\geq$0.9948) across all approaches, indicating robust performance under measurement noise typical in clinical ECG recordings. Also, temporal stability improves dramatically from 0.5304 (Original) to 0.7593 (Aligned Sparse), representing a 43.2\% improvement. This improvement is clinically significant, as it indicates that counterfactuals maintain their validity even when subjected to small temporal shifts common in ECG acquisition.

        The \textit{decision margin} metric shows a trade-off between sparsity and confidence. While the \textit{Original} method achieves a margin of 0.7800, the Aligned Sparse approach shows a decrease to 0.6933. Despite this 11.1\% reduction, the margin remains well above clinical acceptability thresholds, suggesting that the trade-off favoring interpretability and temporal coherence is justified.

    \subsubsection{Class-Specific Performance Analysis}

        \begin{table}[h!]
        \centering
        \caption{Comparison of counterfactual generation methods by metric and target class.}
        \label{tab:stats_by_cf_class}
        \begin{tabular}{llccccc}
        \toprule
        \textbf{Metric} & \textbf{Method} & \textbf{CD} & \textbf{HYP} & \textbf{MI} & \textbf{NORM} & \textbf{STTC} \\
        \midrule
        \multirow{3}{*}{Validity$_{\text{multi}}$} 
         & Original & \textbf{1.000} & \textbf{1.000} & \textbf{1.000} & \textbf{1.000} & \textbf{1.000} \\
         & Sparse & \textbf{1.000} & \textbf{1.000} & \textbf{1.000} & \textbf{1.000} & \textbf{1.000} \\
         & Aligned Sparse & 0.798 & 0.132 & 0.989 & 0.681 & 0.650 \\
        \midrule
        \multirow{3}{*}{$L_0$ sparsity} 
         & Original & 0.985 & 0.982 & 0.982 & 0.987 & 0.984 \\
         & Sparse & 0.810 & \textbf{0.797} & 0.747 & 0.940 & \textbf{0.748} \\
         & Aligned Sparse & \textbf{0.775} & 0.828 & \textbf{0.703} & \textbf{0.914} & 0.817 \\
        \midrule
        \multirow{3}{*}{Noise stability} 
         & Original & \textbf{1.000} & \textbf{1.000} & \textbf{1.000} & \textbf{1.000} & \textbf{1.000} \\
         & Sparse & \textbf{1.000} & \textbf{1.000} & \textbf{1.000} & \textbf{1.000} & \textbf{1.000} \\
         & Aligned Sparse & \textbf{1.000} & 0.978 & \textbf{1.000} & 0.991 & 0.987 \\
        \midrule
        \multirow{3}{*}{Margin} 
         & Original & \textbf{0.607} & \textbf{0.446} & \textbf{0.954} & \textbf{0.695} & 0.598 \\
         & Sparse & 0.605 & 0.445 & 0.946 & 0.691 & 0.597 \\
         & Aligned Sparse & 0.472 & 0.311 & 0.900 & 0.467 & \textbf{0.616} \\
        \bottomrule
        \end{tabular}
        \vspace{0.5em}
        \footnotesize Bold values indicate best performance for each target class-metric combination. For $L_0$ sparsity, lower values are better (sparser counterfactuals). For all other metrics, higher values are better.

        \end{table}

        Table~\ref{tab:stats_by_cf_class} reveals significant performance variations across diagnostic classes, providing insights into the inherent challenges of different conditions. 
        Myocardial Infarction (MI) demonstrates the strongest overall performance, with \textit{validity} of 0.989 and superior sparsity ($L_0$: 0.703) in the \textit{Aligned Sparse} condition. This strong performance aligns with clinical expectations, as MI typically presents with distinct ECG features.

        Conduction Disturbances (CD) show robust \textbf{validity} (0.798) but require more extensive modifications ($L_0$: 0.775), reflecting the global nature of conduction abnormalities that affect timing relationships across multiple ECG leads.

        Hypertrophy (HYP), as the minority class in our dataset (12.15\%), presents the most significant challenge, with validity dropping to only 0.132 in the Aligned Sparse condition. However, the effects of class representation extend beyond HYP alone. Table~\ref{tab:stats_by_cf_class} reveals a broader pattern: classes with fewer effective prototypes tend to produce lower-validity counterfactuals. After single-label filtering, prototype availability varied substantially, and NORM retained 377 samples, while HYP retained only 23. This affected classes that frequently co-occur with other conditions; for instance, STTC commonly appears alongside HYP (6.92\% of recordings, Table~\ref{tab:combination_distribution}), reducing the pool of unambiguous examples for both classes. The validity results reflect this pattern: MI achieves 98.9\% validity with 160 filtered samples, while STTC (231 samples) and NORM (377 samples) show moderate validity (65.0\% and 68.1\% respectively), and HYP (23 samples) shows poor validity (13.2\%). These findings reveal important limitations in generalizability: prototype-based counterfactual methods require sufficient representation not only in the original dataset but also among single-label instances to capture the morphological diversity within each diagnostic category.

        \textit{Decision margin} analysis by class reveals MI counterfactuals maintain the highest confidence (0.9), while HYP shows the lowest (0.311). This pattern correlates strongly with \textit{validity} performance, pointing out that classes with poor validity also suffer from low confidence in successful cases.

\subsection{Advantages of the prototype-based counterfactuals}\label{sec:prototype_advantages}
    In a clinical context, our prototype‐based real-time explanation framework for ECG delivers three concrete benefits:
    \begin{itemize}
        \item \textbf{Computational efficiency:} By clustering rule sets into \(k\) representative prototypes, inference requires only \(k\) prototype comparisons instead of \(N\) individual-instance evaluations, reducing complexity from $\mathcal{O}(N)$ to $\mathcal{O}(k)$, where $k \ll N$. 
        \item \textbf{Clinical validity:} Using actual ECG samples as prototypes guarantees physiologically plausible waveforms, avoiding the synthetic artifacts that can arise from interpolation or generation methods.
        \item \textbf{Interpretability:} Each prototype represents a cluster of similar cases, allowing clinicians to understand not just individual predictions but also the typical patterns within each diagnostic category.
    \end{itemize}

    Medoid-based selection enables our prototypes to capture within-class morphological diversity by choosing the most centrally located ECG in each cluster, ensuring both coverage and representativeness. 
    This balance is crucial for producing concise, clinically meaningful counterfactual exemplars across heterogeneous patient populations. 

    Complementing this representative selection, our sparsity optimization ensures that modifications are restricted to the most influential diagnostic regions, preserving the patient's baseline morphology while allowing the clinician to focus on critical pathological changes.

\subsection{Explanation Representation and Expert Feedback}

    While technical performance metrics provide essential validation of our approach, it is important to distinguish between two complementary notions of counterfactual quality: 
    (i) \textit{validity}, which measures whether the counterfactual successfully changes the model's prediction, and 
    (ii) \textit{clinical plausibility}, which assesses whether the waveform modifications correspond to physiologically meaningful changes that clinicians would recognize. 
    A counterfactual may achieve perfect validity while producing waveform alterations that lack clinical interpretability, or conversely, suggest clinically meaningful changes that do not cross the model's decision boundary. 
    To evaluate clinical plausibility and understand healthcare professionals' cognitive needs, we conducted semi-structured interviews with three practitioners spanning 3 to 40+ years of ECG interpretation experience and varying levels of familiarity with AI systems. 
    Each expert was presented with different visualization modes, including lead-wise importance plots, attention span highlights, and single vs. multiple counterfactual overlays, for identical ECG inputs. 
    They evaluated these modes based on interpretability, cognitive load, clinical relevance, and workflow compatibility.
    Evaluation protocol is documented in Appendix~\ref{secA4}. 

    All experts strongly favored single counterfactual visualizations paired with attention highlighting, citing reduced cognitive effort and improved clarity in identifying diagnostic shifts. Rather than viewing all possible class transitions, they preferred focused, user-driven alternatives relevant to specific clinical scenarios. Regarding clinical plausibility, experts emphasized the importance of aligning lead-level explanations with established pathological mechanisms. For example, when reviewing MI counterfactuals, experts expected modifications concentrated in leads corresponding to affected myocardial regions. They shared concerns when model attributions diverged from these clinically expected patterns, such as when a NORM-to-MI counterfactual modified precordial leads inconsistently with known infarct localization. However, experts also acknowledged cases where model-highlighted regions, while unexpected, revealed subtle waveform features they might have overlooked, suggesting potential for AI-assisted pattern discovery. These observations underscore that validity and clinical plausibility, while related, capture distinct aspects of counterfactual quality.

    One of the experts provided a valuable insight: the granularity of explanation should be adapted to its clinical or research context. 
    In investigative settings, detailed probabilistic confidence scores and channel-weight attributions help uncover novel ECG patterns, while in everyday clinical practice, a simple binary indication of importance proves both sufficient and more user-friendly. 
    The expert further advised tailoring visualization scope by care environment: on general wards, highlighting a single representative beat for non-urgent cases; in emergency departments, offering a concise rationale for deviation from normal to accelerate triage decisions; and in specialized outpatient clinics, presenting richer counterfactual scenarios to aid in distinguishing disease subtypes and supporting complex diagnostic reasoning. 

    These findings underscore the need for interactive, clinician-controlled explanation systems. Experts expressed a clear preference for adaptive visualizations over static, exhaustive outputs and highlighted integration with diagnostic workflows as essential. Their feedback directly informed our decision to prioritize an interactive platform allowing case-specific exploration, real-time adjustment of explanation complexity, and seamless integration with clinical tools.

\subsection{Limitations and Challenges}

    Several limitations emerge from our analysis. 
    The \textit{validity-sparsity trade-off} in the aligned method represents a fundamental challenge of achieving physiologically realistic counterfactuals while maintaining model prediction changes. 
    This trade-off reveals important limitations in generalizability that correlate with class representation. While HYP represents the most extreme case (12.15\% of the dataset, only 23 samples after filtering, 13.2\% validity), the pattern extends to other classes: STTC and NORM achieve only 65.0\% and 68.1\% validity respectively as target classes, despite NORM being the majority class in the original distribution. This apparent paradox is explained by examining the diversity of prototypes and their co-occurrence patterns. Classes that frequently appear in multi-label combinations (Table~\ref{tab:combination_distribution}) lose more samples during single-label filtering, regardless of their original prevalence. 

    The detailed results in Appendix~\ref{appendix:results} (Tables B2–B8) further demonstrate that counterfactual quality degrades in both directions: when generating counterfactuals from minority classes (limited source diversity) and to minority classes (limited prototype options). These findings suggest that prototype-based counterfactual methods have inherent limitations for underrepresented classes in imbalanced, multi-label medical datasets. 
    Future work should investigate targeted sampling strategies, synthetic prototype augmentation, class-aware prototype selection, and relaxed filtering criteria that retain informative multi-label samples to address these disparities.

    Furthermore, the reliance on {R-peak alignment} assumes a definable cardiac cycle, limiting the method's scope. 
    While effective for the morphological and conduction pathologies (MI, HYP, STTC, CD) examined in this study, this approach faces challenges with complex arrhythmias characterized by irregular rhythms (e.g., atrial fibrillation) or absent R-peaks. 
    Future application to arrhythmia-specific tasks would require integrating rhythm-agnostic alignment strategies.

    The \textit{computational complexity} of the alignment process, while acceptable for our current implementation, may limit real-time deployment in resource-constrained clinical environments. 
    The DTW-based alignment requires $O(n^2)$ operations per prototype comparison, which can potentially limit scalability to larger prototype databases.
    Furthermore, our timing measurements ($<1$ second per recording) were obtained on research computing infrastructure and have not been validated on clinical workstations, embedded devices, or higher sampling frequencies (e.g., 500Hz). Scalability testing across diverse deployment environments remains necessary to confirm real-time viability.

    \textit{Evaluation} scope remains limited to the PTB-XL dataset, and generalization to other ECG datasets, different demographics, or varying acquisition protocols requires further validation. 
    Additionally, our expert evaluation involved only three clinicians in semi-structured interviews without a standardized protocol, which limits the generalizability of usability findings. While initial feedback was positive, larger-scale studies with structured evaluation protocols are needed to establish definitive conclusions about clinical utility.

\subsection{Future Improvements}
    The results presented in Section~\ref{subsec:res}, demonstrate significant potential for deployment in clinical ECG interpretation systems. 
    The high validity rates for common conditions (MI, CD) and the improved interpretability through sparse modifications address key requirements for AI-assisted diagnosis. 
    The framework's ability to generate explanations in near real-time makes it suitable for point-of-care applications, while the use of real-patient medoids facilitates clinical verifiability by referencing actual cases. 

    Future improvements should focus on \textit{adaptive validity thresholds} that balance explanation quality with clinical requirements for different conditions. 
    For highly critical conditions, perfect validity may be essential.
    \textit{Hybrid prototype selection strategies} could address class imbalance by combining data-driven clustering with expert-curated examples for rare conditions. 
    This approach would ensure that prototype databases contain representative examples, even for uncommon pathologies. 
    In addition, \textit{real-time adaptation} mechanisms could allow the system to learn from clinician feedback, gradually improving prototype selection and counterfactual generation based on expert preferences and clinical outcomes.

    Our current overlays on ECG plots use SHAP values. 
    To enhance the contrast and readability of ECG feature highlights, we will investigate alternative numeric attribution methods beyond SHAP~\citep{Lundberg2017}, such as LIME~\citep{LIME}, Integrated Gradients~\citep{pmlr-v70-sundararajan17a}, DeepLIFT~\citep{DeepLIFT_2017}, and SmoothGrad~\citep{smilkov2017smoothgrad}. 
    We also plan to explore non-linear transformations of SHAP values or thresholds to enhance contrast and clarity in the visual highlights. 

    The metrics used in our evaluation process capture distinct aspects of counterfactual quality. 
    Similarly to \cite{darias2025evaluating} and \cite{wang2023operationalizing}, the natural next step is to combine them into a single composite quality metric \(Q\) that aggregates four key criteria: validity, sparsity (through the \(L_0\) ratio), stability, and decision margin. 
    ($\textit{Q}(\mathbf{x}') = w_v \cdot \text{validity} + w_s \cdot (1 - \text{sparsity\_ratio}) +  w_{st} \cdot \text{stability\_score} + w_m \cdot \text{decision\_margin}$)
    The main challenge lies in determining the optimal weights \((w_v, w_s, w_{st}, w_m)\), where each weight respectively governs the contribution of each of the aforementioned metrics, for a given clinical application. 
    In future work, we will employ systematic grid search and Bayesian optimization to select weights that best align \(Q\) with clinician-assessed explanation quality. 
    Furthermore, integrating a human(experts)-in-the-loop feedback pipeline will allow iterative, live fine-tuning of these weights based on real-world clinical validation.

    The choice of similarity metric for prototype selection may benefit from further investigation. While DTW accommodates heart rate variability, alternative metrics such as correlation-based measures or frequency-domain distances may better capture clinically meaningful similarities for specific diagnostic contexts, and the optimal choice is likely use-case dependent.

    Beyond algorithmic metrics, translating these results into clinical practice encounters integration hurdles comparable to those in cybersecurity infrastructure protection~\citep{SCHMITT2023100520}. 
    Critical requirements include real-time responsiveness, strict governance, and managing model lifecycle updates amidst changing patient populations. 
    Our framework is designed to meet these demands: sparse, prototype-based explanations are computationally efficient for real-time deployment and provide a transparent audit trail. 
    However, the current implementation represents a foundational algorithmic step. 
    Achieving full clinical utility within a Clinical Decision Support System (CDSS) requires evolving from static outputs to a robust human-in-the-loop workflow. 

    Future work will therefore prioritize developing an \textit{interactive explanation platform} that grants clinicians direct control over the decision support process. 
    Such an interface would allow users to explore different counterfactual scenarios, adjust sparsity levels, and visualize the sensitivity of explanations to various modifications. 
    Furthermore, the platform would enable user-controlled multi-label exploration, allowing clinicians to select specific comorbidity combinations as counterfactual targets when clinically relevant, rather than being limited to single-class transitions. 
    This approach addresses the current limitation of fixed prototype selection while maintaining interpretability through user-driven scope control. 
    A systematic evaluation of these features will require extensive user studies to determine the optimal interaction patterns for complex multi-label scenarios.

    \section{Conclusions}\label{sec:conclusion}
    \label{sec:conc}

Electrocardiogram (ECG) interpretation is essential for diagnosing cardiac disorders under urgent clinical conditions. 
Modern deep learning classifiers achieve remarkable accuracy by extracting complex features from large ECG datasets, but they operate as opaque models (a.k.a. black boxes). 
This lack of transparency undermines clinician confidence and obstructs their adoption in everyday practice, where understanding the reasoning process behind a decision is mandatory. 
Consequently, delivering trustworthy, physiology-based explanations in real time is critical to support rapid, evidence-driven patient care. 
Especially useful in this situation are counterfactual explanations, because they reveal minimal and plausible changes needed to alter ECG classification. 

This work makes five key contributions. 
First, we validate counterfactual and heatmap visualizations with domain experts in ECG interpretation. 
Second, we propose a prototype-guided selection method that uses real ECG recordings as counterfactual templates to achieve physiological validity. 
Third, we develop alignment and sparsification optimization techniques to map and prune prototypes, yielding clear, query-specific explanations. 
Fourth, our approach generalizes beyond ECG to other multivariate time-series domains, such as wearable patient vitals, industrial sensor networks, or predictive maintenance systems, since its core components remain model-agnostic and can be adapted with modest domain-specific alignment. 
Fifth, we adopt evaluation metrics for counterfactual explanations, originally developed in the XAI literature, and tailor them to the unique characteristics of multivariate time-series ECG signals, ensuring meaningful assessment of our method’s fidelity. 

Expert interviews with ECG specialists confirmed that our counterfactual visualizations and heatmaps improve clinical reasoning and align with physiological knowledge. 
We applied a novel~\emph{Post-hoc Attribution Rule extraction (PHAR)} method to extract logical rules for each ECG sample, capturing key morphological features such as amplitude thresholds. 
From these rules, we defined a distance metric that measures similarity between ECG waveforms while preserving physiological plausibility. 
Clustering with medoids yielded representative prototypes from non-query classes, which served as counterfactual examples. 
A dedicated alignment algorithm then mapped each prototype onto the query sample’s temporal structure. 
Prototype sparsification produced more concise, targeted explanations. 
Quantitative evaluation on validity, sparsity, stability, and margin metrics confirmed excellent validity for both \textit{Original} and \textit{Sparse} counterfactuals, with only minor degradation for sparse aligned variants; margins remained clear and stability high across all types.

To conclude, our explainable ECG classification framework unites high accuracy with real-time transparency. 
Clinicians can potentially receive intuitive, counterfactual insights that foster trust and support rapid decision-making. 
Moreover, its largely model-agnostic design extends naturally to diverse multivariate time-series data—enabling applications from wearable patient vitals to industrial sensor monitoring and predictive maintenance.

\bmhead{Acknowledgements}
    The authors thank the Faculty of Physics, Astronomy and Applied Computer Science at Jagiellonian University for computational resources.

\section*{Declarations}
\bmhead{Ethics approval and consent to participate}
    This study involved two components. 
    First, it used public, de-identified ECG datasets. 
    Second, it included semi-structured interviews with domain experts (ECG specialists). 
    All expert participants were adults, participated voluntarily, and provided informed consent before taking part. 
    The activity collected only professional opinions about methods and clinical workflows. 
    No personal data or patient information were collected. 
    Responses were recorded and analysed in anonymized, aggregated form. 
    In accordance with the authors' institutional guidelines, this expert consultation did not require formal approval by a Research Ethics Committee because it did not involve patients, vulnerable individuals, or identifiable personal data. 

\bmhead{Consent for publication}
    All expert participants consented to the publication of anonymized and aggregated results. 
    No identifiable quotations or personal information are presented. 

\bmhead{Availability of data and material}
    This study uses only public, de-identified ECG datasets. 
    The PTB-XL dataset is publicly available from community repositories listed in the References. 
    No new patient data were collected by the authors.

    The code used for data processing, model training, explanation generation, and figure creation is available from the authors upon reasonable request for academic use. 
    Trained model weights and auxiliary artifacts (e.g., SHAP attributions and prototype identifiers) can also be shared upon request.  
    To protect participant anonymity, raw interview notes and individual survey responses are not publicly shared. 
    The interview protocol is available from the authors upon reasonable request for academic research. 

    There are no proprietary restrictions that prevent sharing of the materials. 
    There are no license or contractual restrictions that prevent sharing for academic research.

\bmhead{Competing interests}
    The authors declare no competing interests.

\bmhead{Funding}
    This paper is part of a project that has received funding from the European Union’s Horizon Europe Research and Innovation Programme under Grant Agreement No.\ 101120406. 
    The paper reflects only the authors’ view and the European Commission is not responsible for any use that may be made of the information contained herein. 

    This work was supported by the ELLIIT Excellence Center at Link\"oping–Lund in Information Technology in Sweden. 

    Contribution of Maciej Mozolewski for the research for this publication has been supported by a grant from the Priority Research Area (DigiWorld) under the Mark Kac Center for Complex Systems Research Strategic Programme Excellence Initiative at Jagiellonian University. 

    This publication has been funded by the SFI NorwAI (Centre for Research-based Innovation, 309834). Betül Bayrak gratefully acknowledges the financial support from the Research Council of Norway and the partners of the SFI NorwAI.

\bmhead{Author contributions}
    ~    \vspace{1em}

    \textbf{Maciej Mozolewski:} 
    Prepared and preprocessed the ECG dataset, including data splitting, signal normalization, and background sample preparation for SHAP analysis. 
    Implemented the SHAP attribution pipeline using GradientExplainer with efficient HDF5-based storage. 
    Designed and developed the deep ECG classifier with an advanced training regimen and threshold calibration based on F1 score. 
    Adapted the \emph{PHAR} rule-extraction module for the multiclass setting, enabling overlapping class assignments per instance. 
    Integrated R-peak detection and prototype temporal alignment. 
    Executed computational analyses, collated the resulting metrics and figures, and selected representative ECG signals for visualization. 
    Automated the end-to-end data workflow for comprehensive table and figure generation. 
    Drafted, structured, and unified the Introduction and Background sections of the manuscript. 

    \textbf{Betül Bayrak:} Conceptualization of the counterfactual explanation generation approach. Development of prototype-guided counterfactuals, including implementation of DTW distance computation and clustering, prototype extraction, counterfactual selection, and sparsity optimization method. Development of quantitative evaluation metrics, which are validity, sparsity, stability, and margin metrics, and implementation of the query processing workflow. Organization and execution of cardiologist interviews, survey design, and data collection; creation of workflow and architecture figures.

    \textbf{Maciej Mozolewski and Betül Bayrak:} Conceptualization of the study and manuscript framework; interpretation of results; creation of data visualizations; drafting of Materials and Methods, Results and Discussion, and Conclusions; final review and approval of the manuscript. 

    \textbf{Grzegorz J. Nalepa:} Provided critical feedback on content and structure. Ensured clarity, consistency, and accuracy of language. Edited and refined the final manuscript. 

    \textbf{Kerstin Bach:} Facilitated interviews with ECG experts by identifying suitable experts and making introductions. Advised on interview scope and survey materials. Reviewed the final manuscript.

    \bibliography{sn-bibliography}

    \clearpage
    \begin{appendices}

    \section{Details of model training and threshold selection}\label{secA3}
        \subsection{Training} \label{appendix:training}
    Training optimizes the multi-label binary cross-entropy loss using the \textit{Adam optimizer} with a constant learning rate of $10^{-4}$. The model runs for up to 1000 epochs, with training and validation losses recorded each epoch. 
    Early stopping halts training if the validation loss does not improve for 25 consecutive epochs, restoring the best model state. 
    To maintain stability, gradient norm clipping is applied throughout training.
    \begin{figure}[h]
      \centering
      \includegraphics[width=0.8\linewidth]{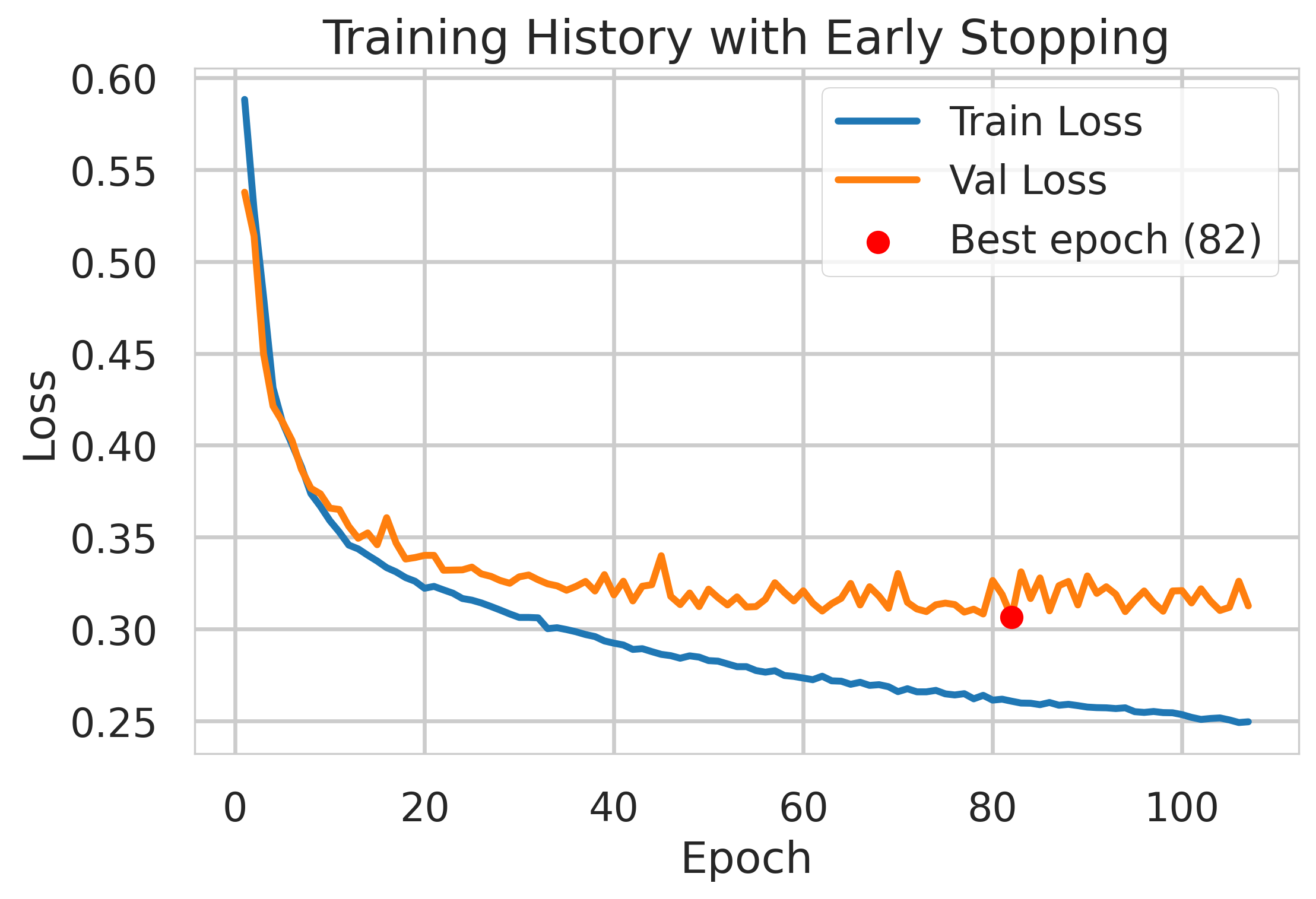}
      \caption{Training and validation loss with early stopping.}
      \label{fig:training-history}
    \end{figure}
    Figure~\ref{fig:training-history} displays the training and validation losses over epochs. 
    Training loss steadily decreases, while validation loss plateaus around epoch 80, at which point early stopping was triggered. 

    Evaluation on the test set using a uniform threshold of 0.5 for all classes yielded a Hamming loss of 0.1285, corresponding to an exact‐match accuracy of 87.15\%. 
    The Jaccard score reached 0.6754, while the F1 metrics were 0.7359 (micro), 0.6973 (macro) and 0.7295 (weighted). 
    No class-specific thresholds were applied in this evaluation.

    To understand where the model still struggles, we conducted an error‐pattern analysis and collected all misclassifications affecting at least 1\% of the test samples (see Table \ref{tab:misclass}). 
    This breakdown reveals, for example, that myocardial infarction cases are most often confused with normal ECGs, and that mixed labels (e.g.\ CD + NORM) sometimes collapse to single‐class predictions. 
    \begin{table}[h]
      \centering
      \caption{Misclassification patterns above 1\% of test samples}
      \label{tab:misclass}
      \begin{tabular}{@{}llr@{}}
        \toprule
        True         & Predicted  & \% of test \\
        \midrule
        MI           & NORM       & 3.14 \\
        NORM         & —          & 2.46 \\
        CD           & NORM       & 1.77 \\
        STTC         & NORM       & 1.55 \\
        CD, NORM     & NORM       & 1.32 \\
        NORM         & STTC       & 1.32 \\
        STTC         & —          & 1.27 \\
        HYP          & NORM       & 1.23 \\
        CD           & CD, MI     & 1.14 \\
        CD, MI       & CD         & 1.05 \\
        NORM         & MI         & 1.00 \\
        \bottomrule
      \end{tabular}
      \vspace{0.5em}

      {\footnotesize
        \textbf{Abbreviations:} 
        NORM – Normal ECG; 
        MI – Myocardial Infarction; 
        STTC – ST/T Change; 
        CD – Conduction Disturbance; 
        HYP – Hypertrophy.
      }
    \end{table}

    \subsection{Threshold selection} \label{appendix:threshold_selection}
    Optimal decision thresholds for each class were found by scanning the validation set probabilities from 0.0 to 1.0 in steps of 0.001 and choosing the value that maximized the binary F1 score for that class. 
    This procedure yielded thresholds of approximately 0.307 for normal ECG (\textit{NORM}), 0.316 for myocardial infarction (\textit{MI}), 0.336 for conduction disturbance (\textit{CD}), 0.352 for ST/T change (\textit{STTC}), and 0.446 for hypertrophy (\textit{HYP}).

    Applying these class‐specific thresholds on the test set produced a Hamming loss of 0.1381 (accuracy 86.19\%), Jaccard score 0.6935, F1 micro 0.7436, F1 macro 0.7094 and F1 weighted 0.7419, showing a slight trade‐off in exact matches but improved overlap and balanced F1 across classes.  

    \begin{figure}[h]
      \centering
      \includegraphics[width=0.8\linewidth]{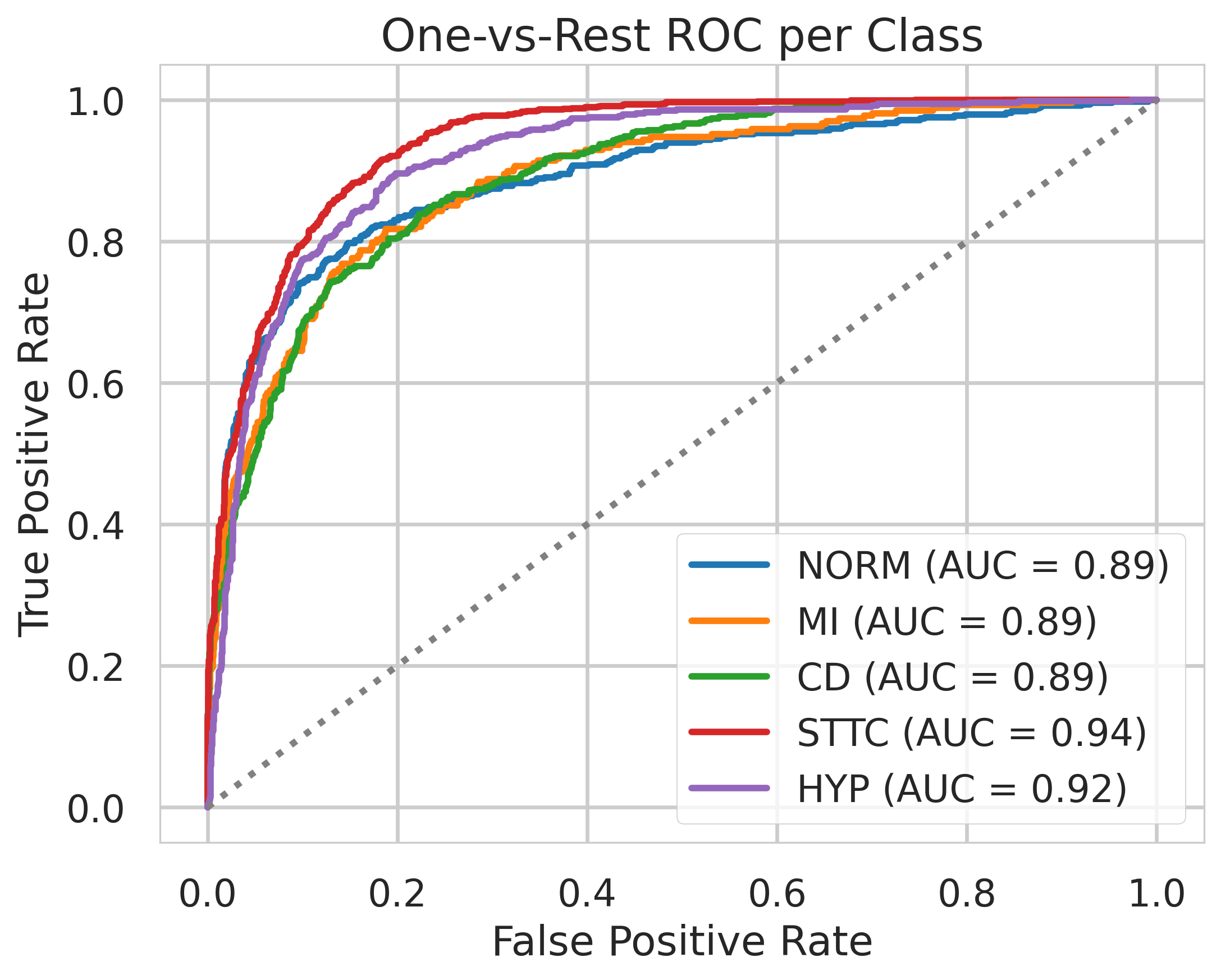}
      \caption{One‐versus‐rest ROC curves per class on validation data.}
      \label{fig:roc-per-class}
    \end{figure}

    Figure~\ref{fig:roc-per-class} plots the true positive rate against the false positive rate for each class as the classification threshold varies. 
    The area under each curve (AUC) quantifies how well the model separates positive from negative examples: ST/T Change (STTC) achieves the highest AUC (~0.94), indicating very good discrimination, followed by Hypertrophy (HYP) at ~0.92. 
    The normal (NORM), infarction (MI) and conduction disturbance (CD) classes all reach AUC~\(\approx0.89\), demonstrating strong but slightly lower separability.

    \subsection{Ablation Study of the Feature Importance Threshold for PHAR} \label{appendix:ablation}
    We performed a sensitivity analysis on a random subset of 350 samples from Fold 10 (the fold dedicated to rule extraction, see Section~\ref{sec:data_preprocessing}) to determine the optimal SHAP feature importance threshold $\tau$ for the {Post-hoc Attribution Rule extraction (PHAR)} rule generation module. 
    It is important to emphasize that at this intermediate stage of the pipeline, the primary objective is not immediate human interpretability, but rather the preservation of the most critical signal information to define distinct clusters.

    \begin{table}[ht]
        \centering
        \caption{Impact of the SHAP selection threshold ($\tau$) on the characteristics of generated rules. 
            The study was conducted on a representative subset of $N=350$ samples ($\approx 16\%$ of the test set). }
        \label{tab:ablation_study}
        \begin{tabular}{c c c}
            \toprule
            \textbf{Threshold ($\tau$)} & \textbf{Feature Count per Rule ($\mu \pm \sigma$)} & \textbf{Inference Time (s)} \\
            \midrule
            45$^{th}$ percentile & $536.8 \pm 458.0$ & 359.0 \\
            50$^{th}$ percentile & $612.1 \pm 422.6$ & 342.6 \\
            70$^{th}$ percentile & $634.4 \pm 415.2$ & 339.5 \\
            \textbf{90$^{th}$ percentile} & $\mathbf{733.1 \pm 284.2}$ & \textbf{336.7} \\
            95$^{th}$ percentile & $489.5 \pm 186.8$ & 332.3 \\
            \bottomrule
        \end{tabular}
    \end{table}

    As shown in Table~\ref{tab:ablation_study} the results highlight two key advantages of the 90$^{th}$ percentile. 
    First, it maximizes pattern granularity: the generated rules contain the highest average number of feature conditions ($733.1$), implying that the algorithm stabilizes a rich set of relevant features rather than discarding them. 
    This high level of detail is crucial for defining homogenous clusters for prototype extraction. 
    Lower thresholds (45--70$^{th}$ percentile) resulted in significantly higher variance in feature counts ($\sigma > 415$). 
    This suggests that forcing the model to consider a vast number of low-importance features (background noise) leads to inconsistent feature selection across samples, where the resulting rules fluctuate between broad and narrow definitions.

    Second, the 90$^{th}$ percentile improves computational efficiency, reducing the average inference time by approximately 22 seconds per sample compared to the 45$^{th}$ percentile baseline, as the search space for perturbations is more constrained to salient features. 
    In contrast, the 95$^{th}$ percentile led to a sharp decrease in feature count, resulting in oversimplified representations that are less effective for subsequent prototype derivation. 
    Thus, $\tau=90$ was selected to balance descriptive richness with computational performance.

    \clearpage

    \section{Metrics for ECG counterfactuals}\label{secA1}
    In this appendix, we report mean values of four counterfactual quality metrics, \textit{multi-class validity} (\(Validity_{multi}\)), \(L_0\) \textit{sparsity}, \textit{noise stability,} and \textit{decision margin} for each initial ECG class. 
For each class, three series are evaluated: the \emph{Original} ECG signal, the \emph{Sparse} counterfactual before alignment, and the \emph{Aligned Sparse} counterfactual after temporal alignment. 
Tables~\ref{tab:stats_query_norm}–\ref{tab:stats_query_hyp} list these metrics for the five single classes (NORM, STTC, CD, MI, HYP), 
while Table~\ref{tab:stats_combined_part1} and Table~\ref{tab:stats_combined_part2} presents the same metrics for two-class combinations. 

\begin{table}[h!]
\caption{Mean counterfactual metrics for initial class NORM (n=564)}
\label{tab:stats_query_norm}
\begin{tabular}{llcccc}
\toprule
Target Class & Series & Validity$_{multi}$ & $L_0$ sparsity & Noise stability & Margin \\
\midrule
\multirow{3}{*}{CD} & Original & 1.0000 & 0.9871 & 1.0000 & 0.6080 \\
 & Sparse & 1.0000 & 0.8114 & 1.0000 & 0.6055 \\
 & Aligned Sparse & 0.8144 & 0.7584 & 1.0000 & 0.4632 \\
\midrule
\multirow{3}{*}{HYP} & Original & 1.0000 & 0.9822 & 1.0000 & 0.4499 \\
 & Sparse & 1.0000 & 0.7892 & 1.0000 & 0.4495 \\
 & Aligned Sparse & 0.1163 & 0.8396 & 1.0000 & 0.2815 \\
\midrule
\multirow{3}{*}{MI} & Original & 1.0000 & 0.9835 & 1.0000 & 0.9591 \\
 & Sparse & 1.0000 & 0.7545 & 1.0000 & 0.9508 \\
 & Aligned Sparse & 0.9929 & 0.6942 & 1.0000 & 0.9062 \\
\midrule
\multirow{3}{*}{STTC} & Original & 1.0000 & 0.9856 & 1.0000 & 0.5932 \\
 & Sparse & 1.0000 & 0.7442 & 1.0000 & 0.5927 \\
 & Aligned Sparse & 0.6162 & 0.8124 & 0.9891 & 0.6176 \\
\bottomrule
\end{tabular}
\end{table}

\begin{table}[h!]
\caption{Mean counterfactual metrics for initial class STTC (n=148)}
\label{tab:stats_query_sttc}
\begin{tabular}{llcccc}
\toprule
Target Class & Series & Validity$_{multi}$ & $L_0$ sparsity & Noise stability & Margin \\
\midrule
\multirow{3}{*}{CD} & Original & 1.0000 & 0.9858 & 1.0000 & 0.6269 \\
 & Sparse & 1.0000 & 0.8136 & 1.0000 & 0.6218 \\
 & Aligned Sparse & 0.6923 & 0.7889 & 1.0000 & 0.5037 \\
\midrule
\multirow{3}{*}{HYP} & Original & 1.0000 & 0.9808 & 1.0000 & 0.4057 \\
 & Sparse & 1.0000 & 0.7485 & 1.0000 & 0.4013 \\
 & Aligned Sparse & 0.0000 & 0.8803 & 0.9000 & 0.4431 \\
\midrule
\multirow{3}{*}{MI} & Original & 1.0000 & 0.9826 & 1.0000 & 0.9404 \\
 & Sparse & 1.0000 & 0.7479 & 1.0000 & 0.9343 \\
 & Aligned Sparse & 1.0000 & 0.7284 & 1.0000 & 0.8790 \\
\midrule
\multirow{3}{*}{NORM} & Original & 1.0000 & 0.9878 & 1.0000 & 0.6840 \\
 & Sparse & 1.0000 & 0.9386 & 1.0000 & 0.6793 \\
 & Aligned Sparse & 0.5733 & 0.9108 & 1.0000 & 0.4134 \\
\bottomrule
\end{tabular}
\end{table}

\begin{table}[h!]
\caption{Mean counterfactual metrics for initial class CD (n=119)}
\label{tab:stats_query_cd}
\begin{tabular}{llcccc}
\toprule
Target Class & Series & Validity$_{multi}$ & $L_0$ sparsity & Noise stability & Margin \\
\midrule
\multirow{3}{*}{HYP} & Original & 1.0000 & 0.9789 & 1.0000 & 0.4410 \\
 & Sparse & 1.0000 & 0.8936 & 1.0000 & 0.4406 \\
 & Aligned Sparse & 0.0000 & 0.9493 & 1.0000 & 0.5537 \\
\midrule
\multirow{3}{*}{MI} & Original & 1.0000 & 0.9775 & 1.0000 & 0.9505 \\
 & Sparse & 1.0000 & 0.6954 & 1.0000 & 0.9427 \\
 & Aligned Sparse & 0.9748 & 0.6762 & 1.0000 & 0.8987 \\
\midrule
\multirow{3}{*}{NORM} & Original & 1.0000 & 0.9881 & 1.0000 & 0.6846 \\
 & Sparse & 1.0000 & 0.9378 & 1.0000 & 0.6791 \\
 & Aligned Sparse & 0.6970 & 0.8784 & 1.0000 & 0.4915 \\
\midrule
\multirow{3}{*}{STTC} & Original & 1.0000 & 0.9845 & 1.0000 & 0.6100 \\
 & Sparse & 1.0000 & 0.7854 & 1.0000 & 0.6079 \\
 & Aligned Sparse & 0.5909 & 0.8513 & 0.9773 & 0.6195 \\
\bottomrule
\end{tabular}
\end{table}

\begin{table}[h!]
\caption{Mean counterfactual metrics for initial class MI (n=111)}
\label{tab:stats_query_mi}
\begin{tabular}{llcccc}
\toprule
Target Class & Series & Validity$_{multi}$ & $L_0$ sparsity & Noise stability & Margin \\
\midrule
\multirow{3}{*}{CD} & Original & 1.0000 & 0.9856 & 1.0000 & 0.6215 \\
 & Sparse & 1.0000 & 0.7767 & 1.0000 & 0.6183 \\
 & Aligned Sparse & 0.8542 & 0.7775 & 1.0000 & 0.4828 \\
\midrule
\multirow{3}{*}{HYP} & Original & 1.0000 & 0.9844 & 1.0000 & 0.4743 \\
 & Sparse & 1.0000 & 0.7944 & 1.0000 & 0.4712 \\
 & Aligned Sparse & 0.2500 & 0.7110 & 1.0000 & 0.2372 \\
\midrule
\multirow{3}{*}{NORM} & Original & 1.0000 & 0.9889 & 1.0000 & 0.6922 \\
 & Sparse & 1.0000 & 0.9374 & 1.0000 & 0.6872 \\
 & Aligned Sparse & 0.7658 & 0.9033 & 0.9910 & 0.4937 \\
\midrule
\multirow{3}{*}{STTC} & Original & 1.0000 & 0.9824 & 1.0000 & 0.5981 \\
 & Sparse & 1.0000 & 0.7454 & 1.0000 & 0.5982 \\
 & Aligned Sparse & 0.7184 & 0.8103 & 0.9854 & 0.6113 \\
\bottomrule
\end{tabular}
\end{table}

\begin{table}[h!]
\caption{Mean counterfactual metrics for initial class HYP (n=21)}
\label{tab:stats_query_hyp}
\begin{tabular}{llcccc}
\toprule
Target Class & Series & Validity$_{multi}$ & $L_0$ sparsity & Noise stability & Margin \\
\midrule
\multirow{3}{*}{CD} & Original & 1.0000 & 0.9826 & 1.0000 & 0.6257 \\
 & Sparse & 1.0000 & 0.8186 & 1.0000 & 0.6287 \\
 & Aligned Sparse & 0.7143 & 0.8076 & 1.0000 & 0.4438 \\
\midrule
\multirow{3}{*}{MI} & Original & 1.0000 & 0.9804 & 1.0000 & 0.9710 \\
 & Sparse & 1.0000 & 0.7297 & 1.0000 & 0.9616 \\
 & Aligned Sparse & 1.0000 & 0.7324 & 1.0000 & 0.9142 \\
\midrule
\multirow{3}{*}{NORM} & Original & 1.0000 & 0.9837 & 1.0000 & 0.7537 \\
 & Sparse & 1.0000 & 0.9426 & 1.0000 & 0.7497 \\
 & Aligned Sparse & 0.8095 & 0.8529 & 1.0000 & 0.5818 \\
\midrule
\multirow{3}{*}{STTC} & Original & 1.0000 & 0.9837 & 1.0000 & 0.6210 \\
 & Sparse & 1.0000 & 0.7516 & 1.0000 & 0.6205 \\
 & Aligned Sparse & 0.7059 & 0.8011 & 1.0000 & 0.6001 \\
\bottomrule
\end{tabular}
\end{table}

\clearpage

\begin{table}[h!]
\caption{Mean counterfactual metrics for combined initial classes. (I)}
\label{tab:stats_combined_part1}
\scriptsize
\setlength{\tabcolsep}{1pt}
\begin{tabular}{lllcccc}
        \toprule
        Initial Class & Target Class & Series & Validity$_{multi}$ & $L_0$ sparsity & Noise stability & Margin \\
        \midrule
        \multirow{9}{*}{\begin{tabular}[t]{@{}l@{}}CD+MI\\\phantom{C}(n=74)\end{tabular}} & \multirow{3}{*}{HYP} & Original & 1.0000 & 0.9812 & 1.0000 & 0.4300 \\
         &  & Sparse & 1.0000 & 0.9080 & 1.0000 & 0.4304 \\
         &  & Aligned Sparse & 0.2000 & 0.7334 & 0.8000 & 0.2321 \\
        \cmidrule{2-7}
         & \multirow{3}{*}{NORM} & Original & 1.0000 & 0.9868 & 1.0000 & 0.6992 \\
         &  & Sparse & 1.0000 & 0.9423 & 1.0000 & 0.6948 \\
         &  & Aligned Sparse & 0.6081 & 0.9375 & 0.9792 & 0.4581 \\
        \cmidrule{2-7}
         & \multirow{3}{*}{STTC} & Original & 1.0000 & 0.9818 & 1.0000 & 0.6040 \\
         &  & Sparse & 1.0000 & 0.7394 & 1.0000 & 0.6011 \\
         &  & Aligned Sparse & 0.6364 & 0.8195 & 0.9884 & 0.6336 \\
        \midrule
        \multirow{9}{*}{\begin{tabular}[t]{@{}l@{}}CD+STTC\\\phantom{C}(n=45)\end{tabular}} & \multirow{3}{*}{HYP} & Original & 1.0000 & 0.9745 & 1.0000 & 0.4135 \\
         &  & Sparse & 1.0000 & 0.8026 & 1.0000 & 0.4113 \\
         &  & Aligned Sparse & 0.0000 & 0.7445 & 1.0000 & 0.3810 \\
        \cmidrule{2-7}
         & \multirow{3}{*}{MI} & Original & 1.0000 & 0.9824 & 1.0000 & 0.9424 \\
         &  & Sparse & 1.0000 & 0.7473 & 1.0000 & 0.9347 \\
         &  & Aligned Sparse & 0.9556 & 0.7090 & 1.0000 & 0.8996 \\
        \cmidrule{2-7}
         & \multirow{3}{*}{NORM} & Original & 1.0000 & 0.9873 & 1.0000 & 0.6883 \\
         &  & Sparse & 1.0000 & 0.9490 & 1.0000 & 0.6852 \\
         &  & Aligned Sparse & 0.6250 & 0.8732 & 1.0000 & 0.4941 \\
        \midrule
        \multirow{9}{*}{\begin{tabular}[t]{@{}l@{}}HYP+STTC\\\phantom{HY}(n=39)\end{tabular}} & \multirow{3}{*}{CD} & Original & 1.0000 & 0.9794 & 1.0000 & 0.5953 \\
         &  & Sparse & 1.0000 & 0.8980 & 1.0000 & 0.5953 \\
         &  & Aligned Sparse & 0.9231 & 0.7974 & 1.0000 & 0.5212 \\
        \cmidrule{2-7}
         & \multirow{3}{*}{MI} & Original & 1.0000 & 0.9746 & 1.0000 & 0.9513 \\
         &  & Sparse & 1.0000 & 0.8007 & 1.0000 & 0.9429 \\
         &  & Aligned Sparse & 1.0000 & 0.7802 & 1.0000 & 0.8942 \\
        \cmidrule{2-7}
         & \multirow{3}{*}{NORM} & Original & 1.0000 & 0.9822 & 1.0000 & 0.7464 \\
         &  & Sparse & 1.0000 & 0.9411 & 1.0000 & 0.7445 \\
         &  & Aligned Sparse & 0.8000 & 0.9234 & 1.0000 & 0.5173 \\
        \midrule
        \multirow{9}{*}{\begin{tabular}[t]{@{}l@{}}MI+STTC\\\phantom{M}(n=31)\end{tabular}} & \multirow{3}{*}{CD} & Original & 1.0000 & 0.9843 & 1.0000 & 0.5927 \\
         &  & Sparse & 1.0000 & 0.8364 & 1.0000 & 0.5929 \\
         &  & Aligned Sparse & 0.6875 & 0.7992 & 1.0000 & 0.4547 \\
        \cmidrule{2-7}
         & \multirow{3}{*}{HYP} & Original & 1.0000 & 0.9851 & 1.0000 & 0.4410 \\
         &  & Sparse & 1.0000 & 0.7961 & 1.0000 & 0.4414 \\
         &  & Aligned Sparse & 0.0000 & 0.8477 & 1.0000 & 0.4405 \\
        \cmidrule{2-7}
         & \multirow{3}{*}{NORM} & Original & 1.0000 & 0.9880 & 1.0000 & 0.6948 \\
         &  & Sparse & 1.0000 & 0.9510 & 1.0000 & 0.6884 \\
         &  & Aligned Sparse & 0.6129 & 0.9172 & 0.9839 & 0.4295 \\
        \bottomrule
    \end{tabular}
\end{table}

\begin{table}[h!]
\caption{Mean counterfactual metrics for combined initial classes. (II)}
\label{tab:stats_combined_part2}
\scriptsize
\setlength{\tabcolsep}{1pt}
\begin{tabular}{lllcccc}
        \toprule
        Initial Class & Target Class & Series & Validity$_{multi}$ & $L_0$ sparsity & Noise stability & Margin \\
        \midrule
        \multirow{9}{*}{\begin{tabular}[t]{@{}l@{}}CD+NORM\\\phantom{C}(n=25)\end{tabular}} & \multirow{3}{*}{HYP} & Original & 1.0000 & 0.9862 & 1.0000 & 0.4743 \\
         &  & Sparse & 1.0000 & 0.7297 & 1.0000 & 0.4681 \\
         &  & Aligned Sparse & 0.5000 & 0.9136 & 1.0000 & 0.3267 \\
        \cmidrule{2-7}
         & \multirow{3}{*}{MI} & Original & 1.0000 & 0.9814 & 1.0000 & 0.9799 \\
         &  & Sparse & 1.0000 & 0.7301 & 1.0000 & 0.9710 \\
         &  & Aligned Sparse & 1.0000 & 0.6587 & 1.0000 & 0.9362 \\
        \cmidrule{2-7}
         & \multirow{3}{*}{STTC} & Original & 1.0000 & 0.9860 & 1.0000 & 0.5683 \\
         &  & Sparse & 1.0000 & 0.8183 & 1.0000 & 0.5718 \\
         &  & Aligned Sparse & 0.5455 & 0.7837 & 1.0000 & 0.6298 \\
        \midrule
        \multirow{9}{*}{\begin{tabular}[t]{@{}l@{}}CD+HYP\\\phantom{C}(n=11)\end{tabular}} & \multirow{3}{*}{MI} & Original & 1.0000 & 0.9780 & 1.0000 & 0.9559 \\
         &  & Sparse & 1.0000 & 0.6796 & 1.0000 & 0.9446 \\
         &  & Aligned Sparse & 1.0000 & 0.7296 & 1.0000 & 0.9022 \\
        \cmidrule{2-7}
         & \multirow{3}{*}{NORM} & Original & 1.0000 & 0.9871 & 1.0000 & 0.6395 \\
         &  & Sparse & 1.0000 & 0.9414 & 1.0000 & 0.6381 \\
         &  & Aligned Sparse & 0.8182 & 0.9621 & 1.0000 & 0.4175 \\
        \cmidrule{2-7}
         & \multirow{3}{*}{STTC} & Original & 1.0000 & 0.9840 & 1.0000 & 0.6257 \\
         &  & Sparse & 1.0000 & 0.7686 & 1.0000 & 0.6271 \\
         &  & Aligned Sparse & 0.6667 & 0.8813 & 1.0000 & 0.6440 \\
        \midrule
        \multirow{9}{*}{\begin{tabular}[t]{@{}l@{}}HYP+MI\\\phantom{HY}(n=8)\end{tabular}} & \multirow{3}{*}{CD} & Original & 1.0000 & 0.9830 & 1.0000 & 0.5671 \\
         &  & Sparse & 1.0000 & 0.8913 & 1.0000 & 0.5725 \\
         &  & Aligned Sparse & 1.0000 & 0.6807 & 1.0000 & 0.4500 \\
        \cmidrule{2-7}
         & \multirow{3}{*}{NORM} & Original & 1.0000 & 0.9873 & 1.0000 & 0.6380 \\
         &  & Sparse & 1.0000 & 0.9301 & 1.0000 & 0.6402 \\
         &  & Aligned Sparse & 0.7500 & 0.9688 & 1.0000 & 0.3957 \\
        \cmidrule{2-7}
         & \multirow{3}{*}{STTC} & Original & 1.0000 & 0.9754 & 1.0000 & 0.5286 \\
         &  & Sparse & 1.0000 & 0.7159 & 1.0000 & 0.5265 \\
         &  & Aligned Sparse & 0.6250 & 0.8909 & 0.8750 & 0.6000 \\
        \midrule
        \multirow{3}{*}{\begin{tabular}[t]{@{}l@{}}NORM+STTC\\\phantom{NOR}(n=4)\end{tabular}} & \multirow{3}{*}{MI} & Original & 1.0000 & 0.9870 & 1.0000 & 0.9424 \\
         &  & Sparse & 1.0000 & 0.7188 & 1.0000 & 0.9361 \\
         &  & Aligned Sparse & 1.0000 & 0.7317 & 1.0000 & 0.8757 \\
        \bottomrule
    \end{tabular}
\end{table}

    \clearpage

    \section{Visualized instances of ECG counterfactuals from the proposed method}\label{secA2}
    \label{appendix:results}
\begin{figure}[h!]
        \centering
        \includegraphics[width=0.78\textwidth]{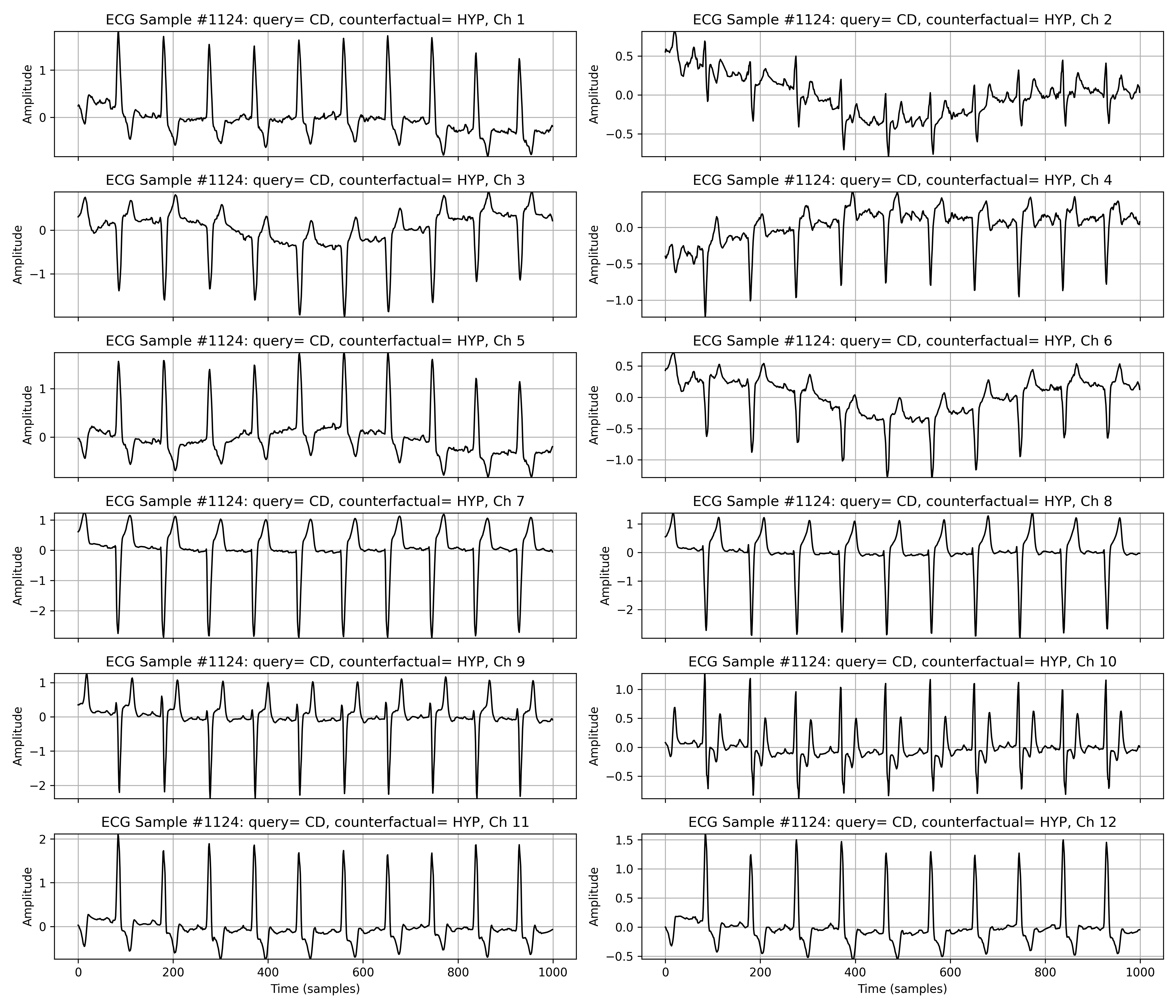}
        \caption{ECG recording from test set (ID 1124) classified as Conduction Disturbance (CD); all 12 leads shown. (\textit{Query})}
        \label{fig:orig}  
    \end{figure}

\begin{figure}[h!]
        \centering
        \includegraphics[width=0.78\textwidth]{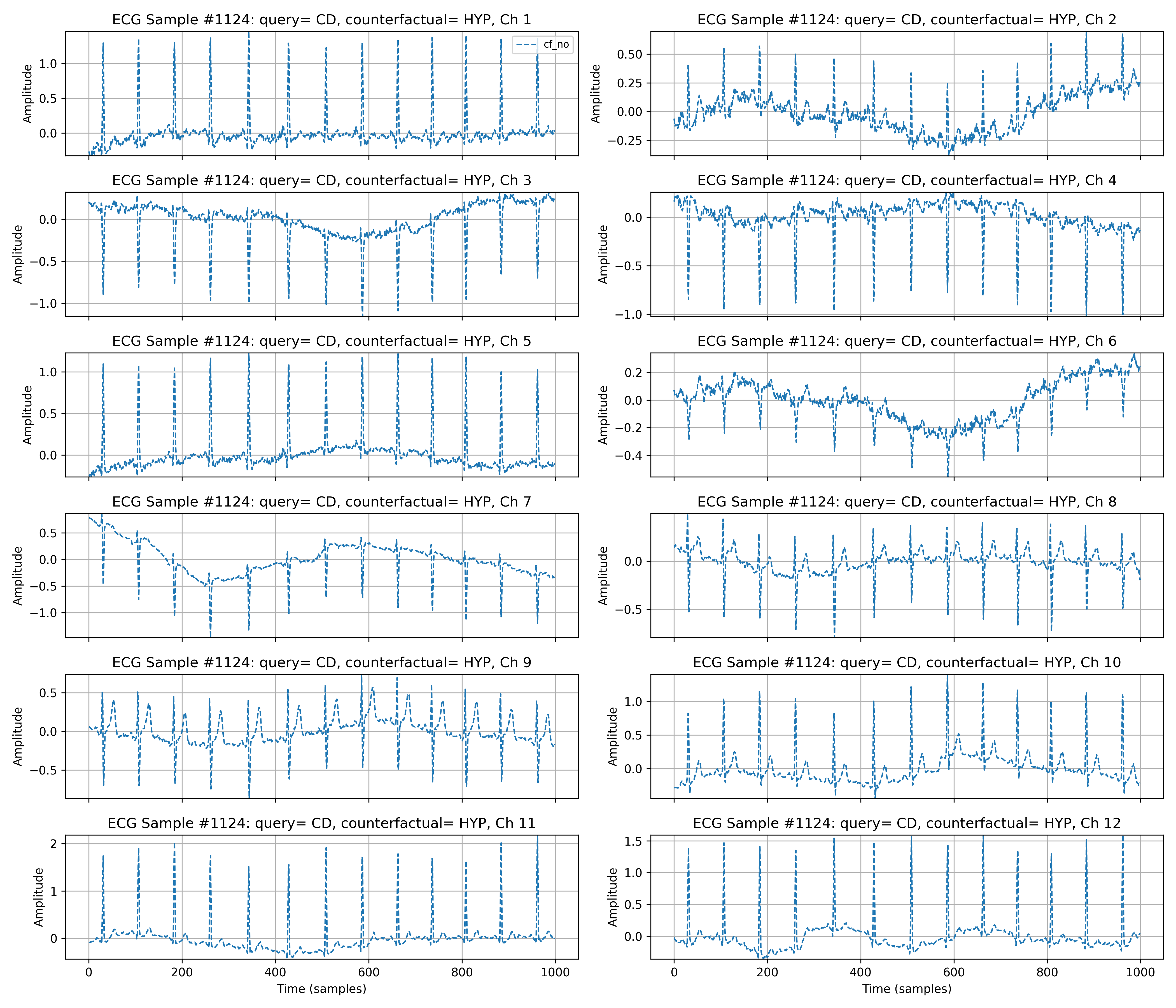}
            \caption{Prototype counterfactual for Hypertrophy (HYP). (\textit{Original})}
        \label{fig:proto_only}  
    \end{figure}

\begin{figure}[h!]
        \centering
        \includegraphics[width=0.78\textwidth]{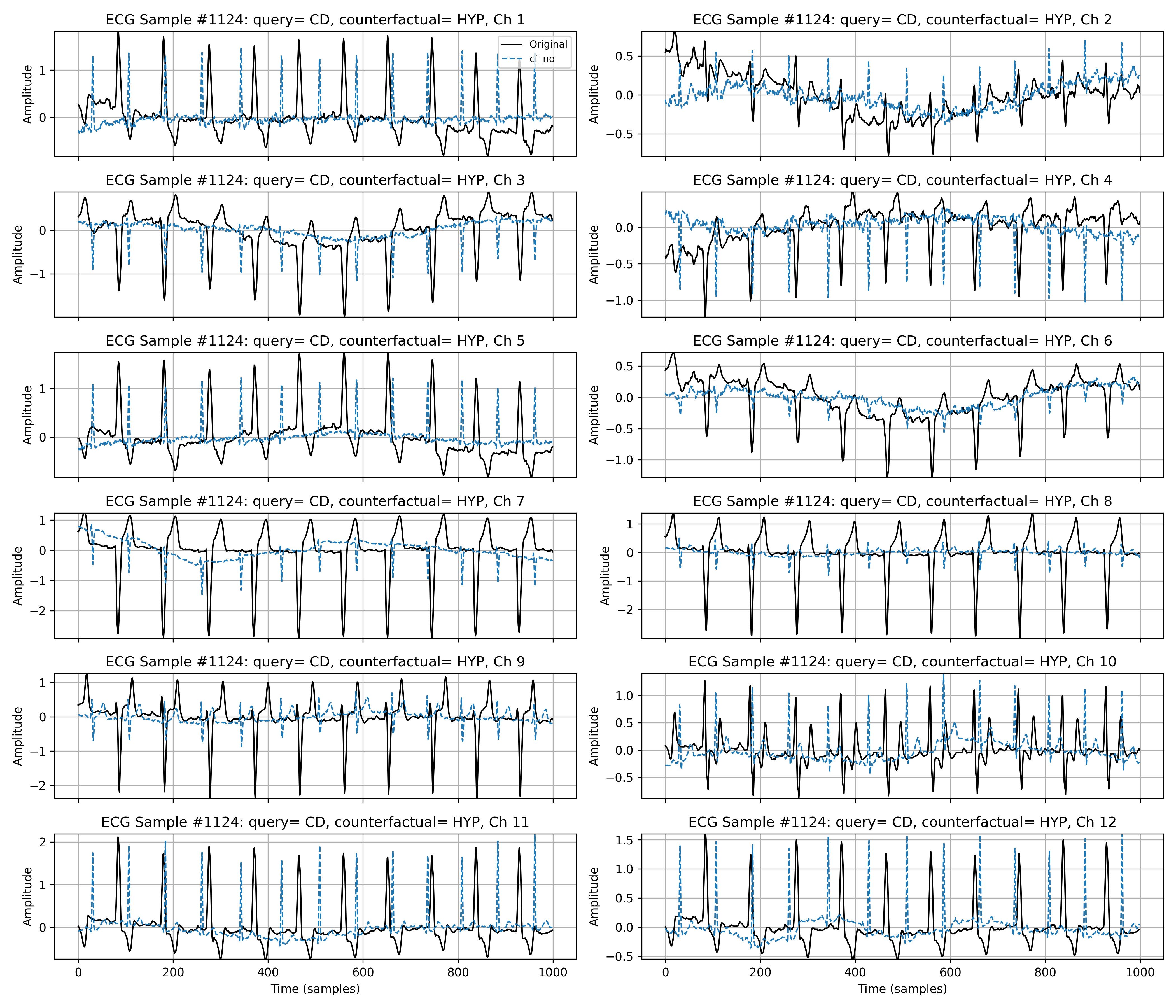}
            \caption{Overlay of the ECG test sample and prototype. (\textit{Query} and \textit{Original})}
        \label{fig:overlay}  
    \end{figure}

\begin{figure}[h!]
        \centering
        \includegraphics[width=0.78\textwidth]{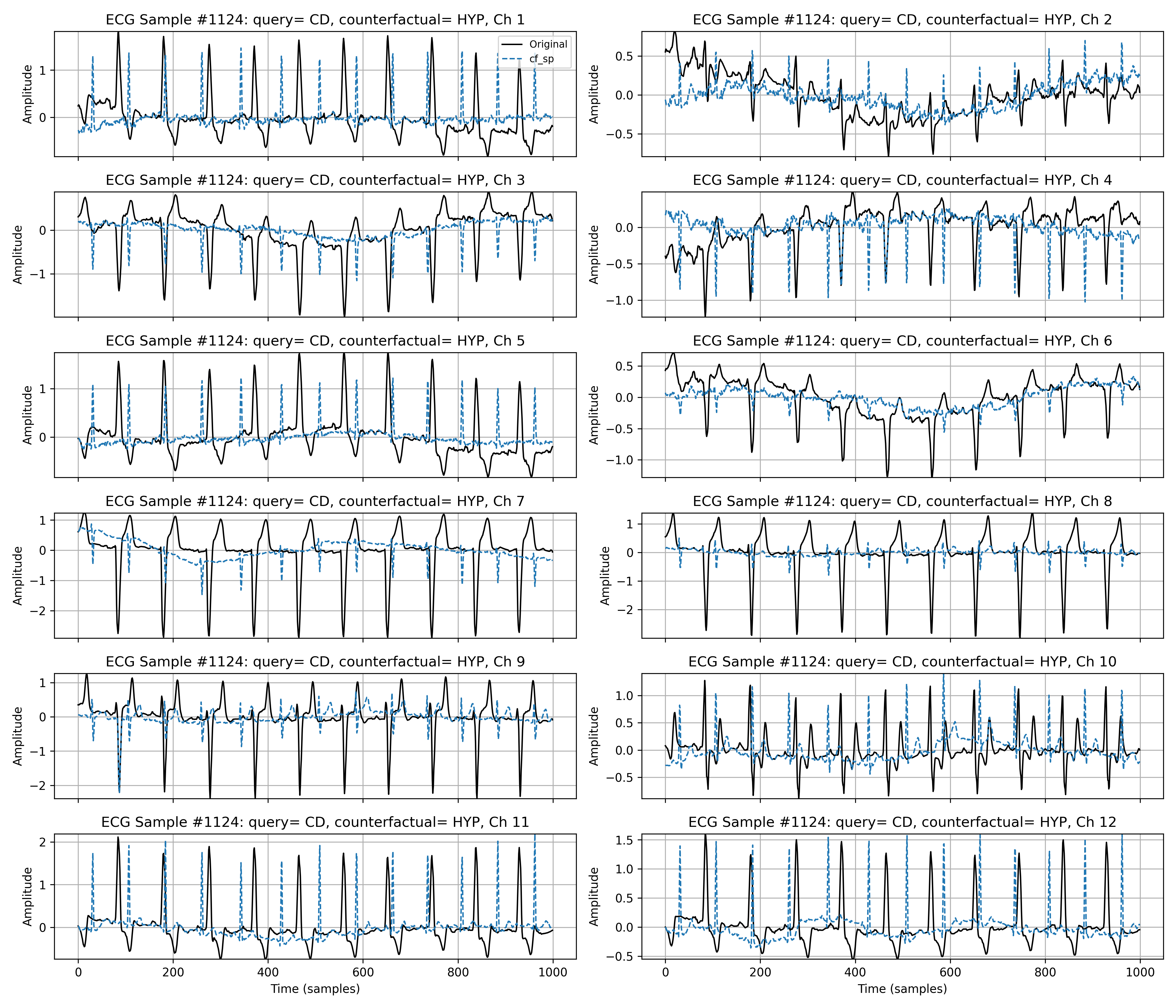}
            \caption{Overlay of the ECG test sample and sparsified prototype, which highlights minimal feature changes. (\textit{Query} and \textit{Sparse})}
        \label{fig:overlay_sparse}  
    \end{figure}

\begin{figure}[h!]
        \centering
        \includegraphics[width=0.78\textwidth]{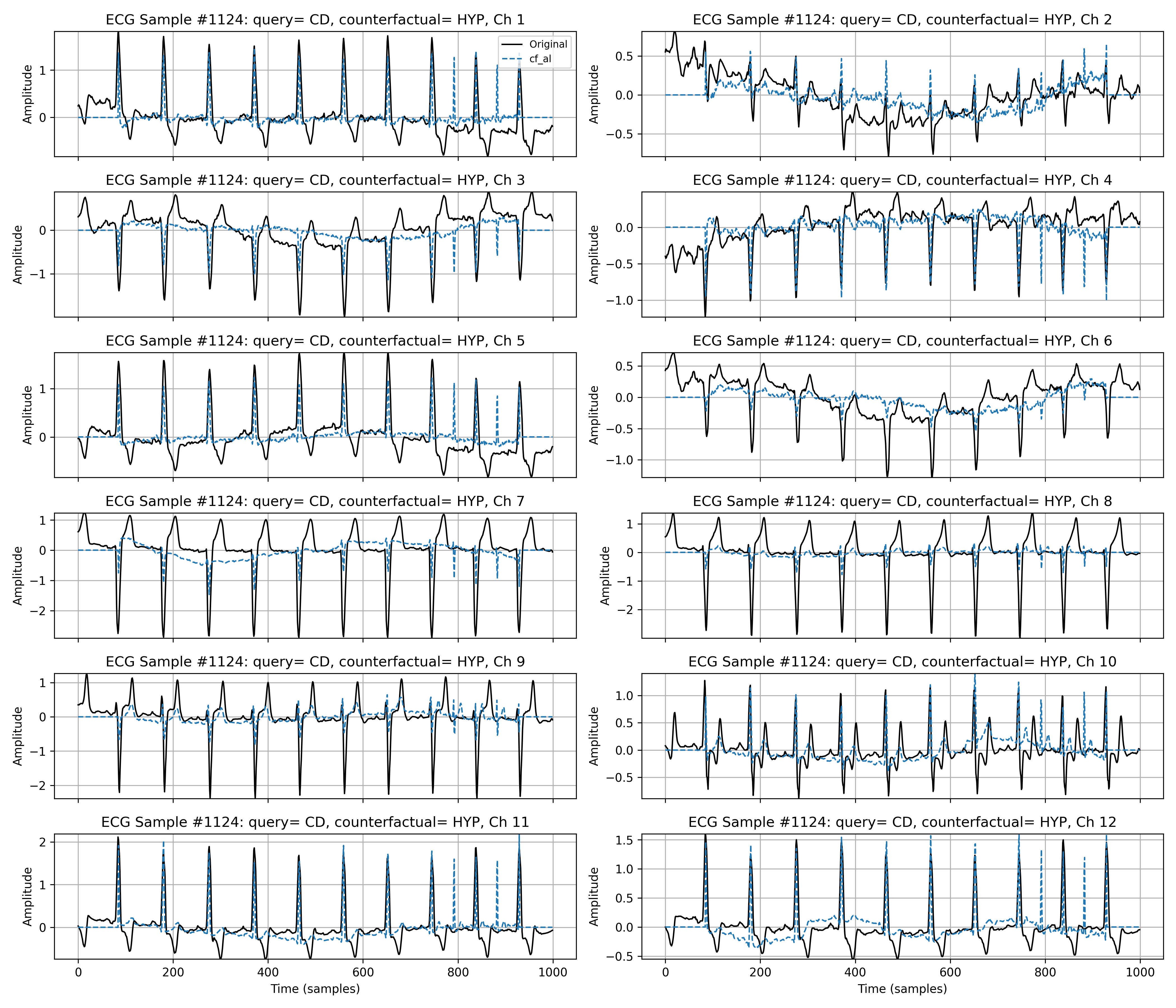}
            \caption{Overlay of the ECG test sample and temporally aligned prototype. (\textit{Query} and \textit{Aligned})}
        \label{fig:overlay_aligned}  
    \end{figure}

\begin{figure}[h!]
        \centering
        \includegraphics[width=0.78\textwidth]{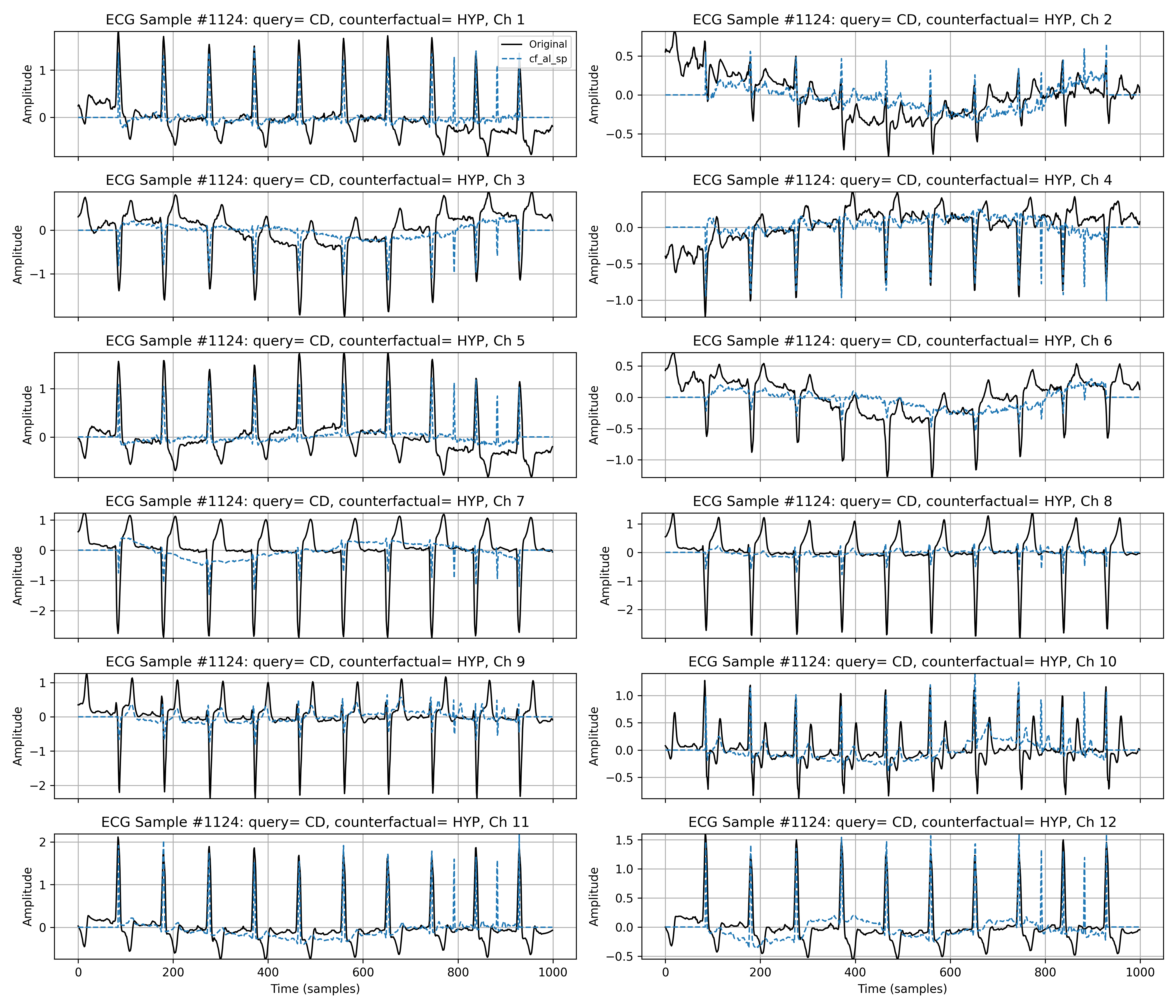}
            \caption{Overlay of the ECG test sample and first aligned then sparsified prototype; minimal, physiologically plausible changes highlighted. (\textit{Query} and \textit{Sparse-Aligned})}
        \label{fig:overlay_aligned_sparse}  
    \end{figure}

\begin{figure}[h!]
        \centering
        \includegraphics[width=0.78\textwidth]{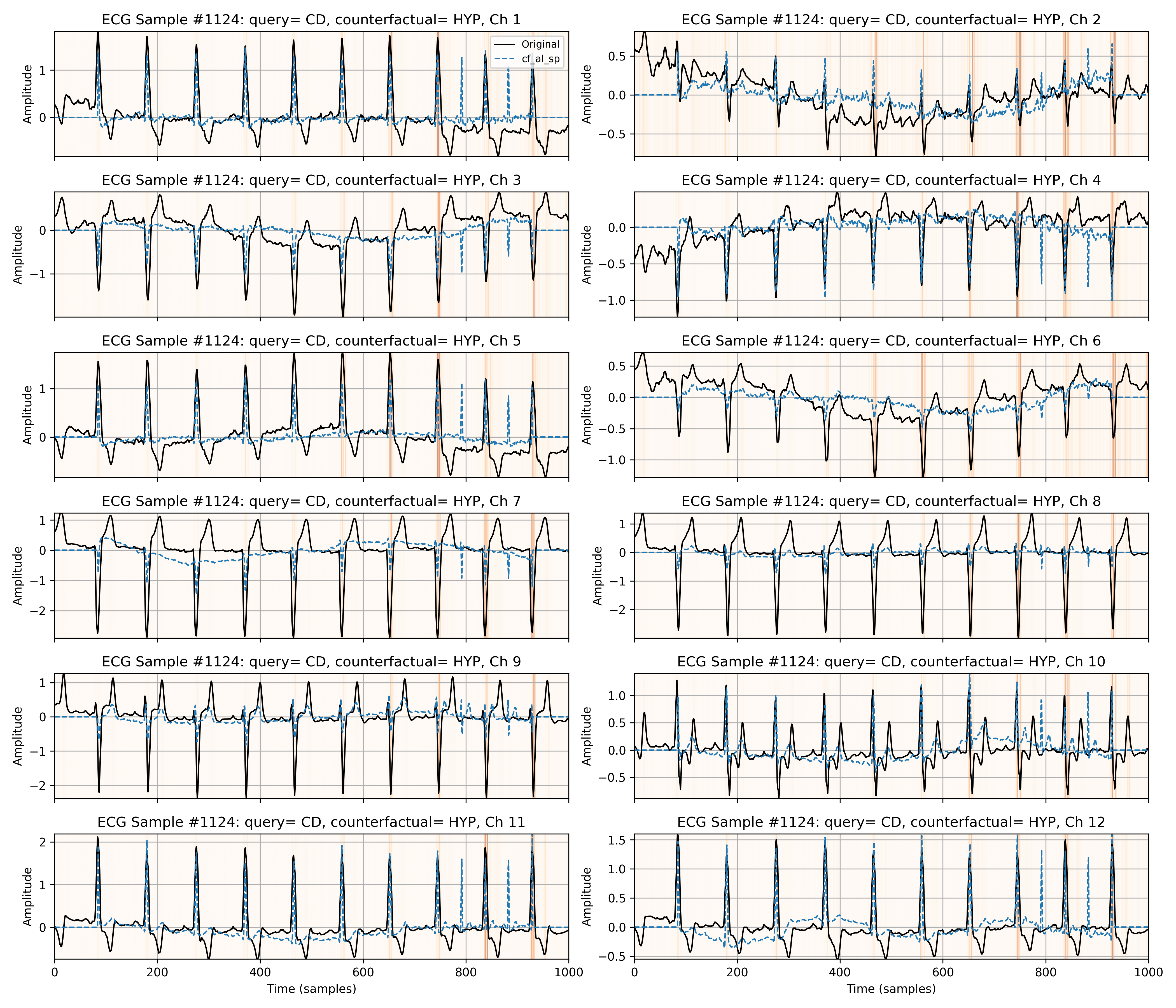}
            \caption{Overlay of the ECG test sample and first aligned then sparsified prototype with SHAP heatmap superimposed. (\textit{Query} and \textit{Sparse-Aligned})}
        \label{fig:overlay_shap}  
    \end{figure}

    This appendix presents visual illustrations of counterfactual generation for ECG sample 1124. 
    The ECG recording from test set (\textit{Query}) classified as Conduction Disturbance (CD) appears in Figure~\ref{fig:orig}, while Figure~\ref{fig:proto_only} shows the Hypertrophy (HYP) prototype (\textit{Original}), which serves as an initial counterfactual by selecting the most representative sample from the HYP cluster, which is also similar to the given ECG sample. 
    For this example, the counterfactual achieves perfect validity (\textit{validity:} 1), successfully flipping the classifier’s prediction from CD to HYP. 
    Subsequent overlays (Figures~\ref{fig:overlay},~\ref{fig:overlay_sparse}) illustrate how the sparsified prototype (\textit{Sparse}) highlights waveform adjustments required to effect the flip, and Figure~\ref{fig:overlay_aligned} demonstrates adjustments with R-peak–based temporal alignment (\textit{Alignment}). 
    The combined alignment and sparsification step is shown in Figure~\ref{fig:overlay_aligned_sparse} (\textit{Sparse-Aligned}), and finally, Figure~\ref{fig:overlay_shap} overlays SHAP-derived attributions on the aligned, sparsified prototype to reveal which ECG features most strongly influence the model’s decision. 


    \clearpage

    \section{Expert Evaluation Protocol and Visual Stimuli} \label{secA4}
    To better understand expert needs and preferences regarding the representation of ECG explanations, we conducted preliminary interviews with medical practitioners. 
These interviews aimed to gain domain insights and learn about clinical requirements for developing suitable decision-support systems. 
Specifically, we focused on expert preferences for different visualization and representation modes, which will inform our future work on explanation delivery.

\subsection{Participant Profiles and Metadata}
For each participant, we recorded professional metadata to contextualize their feedback. The following attributes were collected:
\begin{itemize}
    \item \textbf{Expertise:} Role, specialty, and years of clinical experience.
    \item \textbf{Familiarity (1--5 scale):} Self-reported proficiency in ECG interpretation, Artificial Intelligence (AI), and Clinical Decision Support Systems (CDSS).
    \item \textbf{Format:} Interview setting (in-person or online).
\end{itemize}

\subsection{Evaluation Procedure}
The evaluation was divided into three main phases, utilizing visual stimuli from the PTB-XL dataset:
\begin{enumerate}
    \item \textbf{Baseline Assessment:} Introduction to the counterfactual paradigm and the concept of sparse modifications.
    \item \textbf{Comparative Analysis:} Experts reviewed specific diagnostic transitions to evaluate the model's logic. Key scenarios included:
    \begin{itemize}
        \item Conduction Disturbance (CD) $\rightarrow$ Normal
        \item Conduction Disturbance (CD) $\rightarrow$ Myocardial Infarction (MI)
        \item Hypertrophy (HYP) $\rightarrow$ Normal
    \end{itemize}
    \item \textbf{Visual Representation Review:} Assessment of three visual layers: original query, SHAP-based attention spans (heatmaps), and the final sparse counterfactual overlay.
\end{enumerate}

\subsection{Structured Questionnaire}
Participants provided qualitative and quantitative feedback based on the following structured questions:
\begin{itemize}
    \item \textbf{Lead-Level Contribution:} Is receiving information about individual lead (channel) contributions helpful for diagnosis?
    \item \textbf{Attention Spans:} How do you perceive the integration of SHAP-based heatmaps with counterfactual modifications?
    \item \textbf{Information Density:} Do you prefer a single minimal explanation or a diverse set of counterfactual alternatives?
    \item \textbf{Visual Elements:} Feedback on color coding, transparency, and the clarity of highlighted waveform alterations.
    \item \textbf{Clinical Utility:} Perceived usefulness of the explanations in supporting differential diagnosis and ensuring a seamless workflow fit.
\end{itemize}
    \clearpage

\end{appendices}

\end{document}